%% file: main.tex
\documentclass[runningheads]{llncs}

\usepackage[mobile,year=2024,ID=8999]{eccv}

\usepackage{eccvabbrv}

\usepackage{graphicx}
\usepackage{booktabs}
\usepackage{xcolor,colortbl}
\usepackage{multirow}
\usepackage{amssymb}
\usepackage{pifont}
\usepackage[accsupp]{axessibility}  

\usepackage{makecell}
\usepackage{svg}
\usepackage{xcolor}
\usepackage{subcaption}
\usepackage{wrapfig}
\newcommand{\hlrow}{\rowcolor{black!6}}
\usepackage{algorithm}
\usepackage{algpseudocode}
\usepackage{tablefootnote}

\usepackage[pagebackref,breaklinks,colorlinks,citecolor=eccvblue]{hyperref}

\usepackage{orcidlink}
\newcommand{\encoder}{\mathcal{E}}
\newcommand{\decoder}{\mathcal{D}}

\definecolor{yzybest}{rgb}{0.96, 0.57, 0.58}
\definecolor{yzysecond}{rgb}{0.98, 0.78, 0.57}
\definecolor{yzythird}{rgb}{1.0, 1.0, 0.56}
\definecolor{myblue}{HTML}{1f77b4}
\definecolor{myorange2}{HTML}{ff7f03}
\definecolor{myred}{HTML}{FF0100}
\definecolor{gscolor}{rgb}{1.0,0.6,0.0} %

\definecolor{mygray}{gray}{0.9}
\newcommand*\samethanks[1][\value{footnote}]{\footnotemark[#1]}

\begin{document}

\title{MoDiTalker: Motion-Disentangled Diffusion Model for High-Fidelity Talking Head Generation}
\titlerunning{MoDiTalker}

\author{
Seyeon Kim \inst{1}\thanks{Equal contribution} \and
Siyoon Jin \inst{1}\samethanks \and
Jihye Park \inst{1}\samethanks \vspace{1pt} \and \\
Kihong Kim \inst{2}\and
Jiyoung Kim \inst{1}\and
Jisu Nam \inst{1}\and
Seungryong Kim \inst{1}
}

\authorrunning{S. Kim et al.}

\institute{Korea University, Seoul, Korea \\
\and VIVE STUDIOS, Seoul, Korea \\
\url{https://ku-cvlab.github.io/MoDiTalker}
}

\maketitle

\input{assets/figures/qual_teaser_arxiv}
\vspace{-15pt}
\input{assets/sec/0_abstract}   
\input{assets/sec/1_Introduction} 
\input{assets/sec/2_Related_Work}   
\input{assets/sec/3_Methodology}   
\input{assets/sec/4_Experiments}   
\input{assets/sec/5_Conclusion}


\bibliographystyle{splncs04}
\bibliography{references}

\input{supple_assets/6_supp}

\end{document}

%% file: assets/figures/qual_teaser_arxiv.tex
\begin{center}
    \centering
    \captionsetup{type=figure}
    \includegraphics[width=1\textwidth]{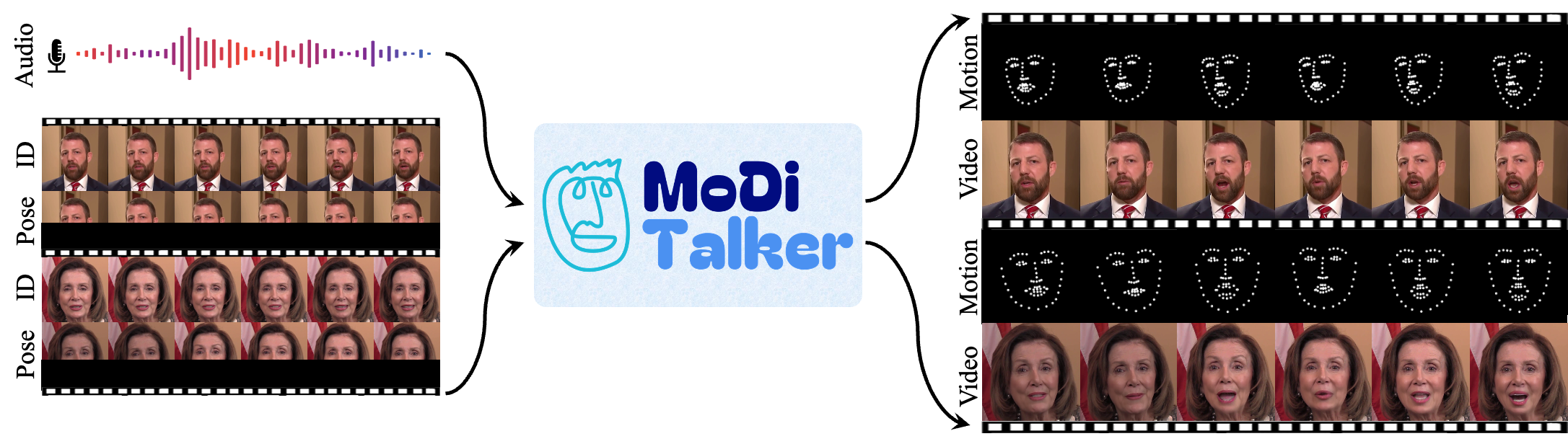}
    \vspace{-15pt}
    \captionof{figure}{We present the \textbf{Mo}tion-\textbf{Di}sentangled diffusion model for high-fidelity \textbf{Talk}ing head gen\textbf{er}ation, dubbed \textbf{MoDiTalker}. This framework generates high-quality talking head videos through a novel two-stage, motion-disentangled diffusion models.}
    \label{fig:teaser}
\vspace{2pt}
\end{center}%

%% file: assets/sec/0_abstract.tex
\begin{abstract}

Conventional GAN-based models for talking head generation often suffer from limited quality and unstable training. Recent approaches based on diffusion models aimed to address these limitations and improve fidelity. However, they still face challenges, including extensive sampling times and difficulties in maintaining temporal consistency due to the high stochasticity of diffusion models. To overcome these challenges, we propose a novel motion-disentangled diffusion model for high-quality talking head generation, dubbed \textbf{MoDiTalker}. We introduce the two modules: audio-to-motion (AToM), designed to generate a synchronized lip motion from audio, and motion-to-video (MToV), designed to produce high-quality head video following the generated motion. 
AToM excels in capturing subtle lip movements by leveraging an audio attention mechanism. In addition, MToV enhances temporal consistency by leveraging an efficient tri-plane representation. Our experiments conducted on standard benchmarks demonstrate that our model achieves superior performance compared to existing models. We also provide comprehensive ablation studies and user study results.

\end{abstract}

%% file: assets/sec/1_Introduction.tex
\section{Introduction}
Audio-driven talking head generation~\cite{suwajanakorn2017synthesizing, zhou2021pose, zhang2023sadtalker, wang2021audio2head, ji2021audio} aims to generate high-fidelity talking head videos with lip movements synchronized to given audio input. This task has been extensively studied for many practical applications~\cite{prajwal2020lip, wang2021one, he2023gaia}, including film production~\cite{prajwal2020lip}, video conferencing~\cite{wang2021one} and digital avatars~\cite{he2023gaia}. A significant challenge in this field is mapping the low-dimensional audio inputs into higher-dimensional lip movements.

To address this, conventional generative adversarial networks (GANs)-based methods~\cite{kr2019towards, prajwal2020lip, zhou2020makelttalk, zhang2023sadtalker, yin2022styleheat, wang2021audio2head, wang2021oneshot} transform the audio embeddings into intermediate facial representations, such as dense motion fields~\cite{yin2022styleheat, zhang2023sadtalker} or 2D/3D facial landmarks~\cite{zhou2020makelttalk, wang2021audio2head, wang2021oneshot}. Notably, recent works~\cite{wang2021audio2head, wang2021oneshot, zhang2023sadtalker} have shown remarkable results by utilizing 3D facial models~\cite{blanz2023morphable} to extract facial landmarks, which are known to effectively capture realistic facial motion. Despite their promising results, however, conventional GAN-based approaches still inherit limitations of GAN, such as training instability~\cite{brock2018large, miyato2018spectral} or mode collapse~\cite{thanh2020catastrophic, bang2021mggan}.

On the other hand, denoising diffusion models~\cite{ho2020denoising, song2020denoising} have recently shown superior performance in image generation tasks, offering more stable training and enhanced fidelity compared to GANs. Inspired by these advancements, some recent studies~\cite{shen2023difftalk, stypułkowski2023diffused, ma2023dreamtalk} propose diffusion-based frameworks for talking head generation. However, these methods often fail to generate realistic facial movements from audio inputs, because they globally incorporate audio embeddings into the diffusion U-Net, which is insufficient to capture local lip movements. Moreover, prior works often rely on a frame-by-frame video generation manner, lagging behind in temporal consistency and inference time complexity compared to GAN-based models~\cite{prajwal2020lip, wang2021one}.

In this paper, we introduce a novel diffusion-based framework for talking head generation, \textbf{MoDiTalker}, which explicitly disentangles the generation process into two distinct stages. First, we present \textbf{A}udio-\textbf{To}-\textbf{M}otion (\textbf{AToM}), a transformer-based diffusion model that generates facial motion sequences conditioned on audio input. Specifically, AToM is designed to predict motion differences from a reference facial landmark for each frame in the sequence. To enhance lip synchronization, we leverage attention mechanisms to distinguish between lip-related and lip-unrelated regions. Second, we propose \textbf{M}otion-\textbf{To}-\textbf{V}ideo (\textbf{MToV}), a motion-conditional video diffusion model that generates temporally consistent talking head videos following facial motion sequences from AToM. Specifically, we leverage tri-plane representations to efficiently condition the video diffusion model. Combining these, MoDiTalker enables high-fidelity talking head generation with enhanced temporal consistency and substantially reduced inference time complexity compared to existing works.

In experiments, our framework achieves state-of-the-art performance on HDTF dataset~\cite{zhang2021flow} surpassing previous GAN-based~\cite{prajwal2020lip,zhou2021pose, wang2021audio2head, zhou2020makelttalk} and diffusion-based~\cite{stypułkowski2023diffused, ma2023dreamtalk, stypulkowski2024diffused} approaches by a large margin. Extensive ablation studies validate our design choices and the effectiveness of our approach.

%% file: assets/sec/2_Related_Work.tex
\section{Related Work}
\subsubsection{GAN-based talking head generation.} 
Early methods~\cite{prajwal2020lip, liang2022expressive, zhou2021pose} in audio-driven talking head generation typically relied on GANs~\cite{goodfellow2020generative}. For instance, Wav2Lip~\cite{prajwal2020lip} leverages a pre-trained SyncNet~\cite{chung2017out} as a lip sync expert to enhance lip synchronization quality. However, since Wav2Lip~\cite{prajwal2020lip} inpaints the generated mouth region, it leads to blurry artifacts and blending issues. Moreover, facial expressions and head poses are overlooked. One of the main challenges in generating talking faces lies in manipulating visual dynamics, which are difficult to directly extract from the given audio inputs. To address this, several methods attempt to exploit 2D~\cite{kumar2017obamanet, suwajanakorn2017synthesizing, song2021everything, zakharov2019few, lu2021live, chen2019hierarchical, zhou2020makelttalk} or 3D~\cite{zhang2021flow, ren2021pirenderer, zhang2023sadtalker} intermediate representations. For example, recent models~\cite{zhang2021flow, ren2021pirenderer, zhang2023sadtalker} utilize 3D Morphable Model~\cite{booth20163d} parameters, including expression, pose, and identity. While these methods can produce smoother and natural movements, they also unavoidably inherit the drawbacks of GANs, such as mode collapse~\cite{thanh2020catastrophic, bang2021mggan} or unstable training~\cite{brock2018large, miyato2018spectral}.

\vspace{-6pt}
\subsubsection{Diffusion-based talking head generation.}
To overcome the inherent challenges of GANs~\cite{thanh2020catastrophic, bang2021mggan, brock2018large, miyato2018spectral}, recent studies~\cite{stypułkowski2023diffused, shen2023difftalk} have proposed diffusion-based frameworks for talking head generation. However, these studies~\cite{stypułkowski2023diffused, shen2023difftalk} often struggle to generate natural facial motion since they globally condition audio embeddings through cross-attention modules or time embedding layers of the denoising U-Net. Furthermore, they rely on frame-by-frame video generation, resulting in significantly slow sampling times and temporal inconsistencies. DreamTalk~\cite{ma2023dreamtalk} generates motions frames with diffusion model but synthesizes the frames with previous GAN-based model~\cite{ren2021pirenderer}. In this paper, we introduce a diffusion model for talking head generation, which leverages intermediate facial representation with minimal computational demand.

\vspace{-6pt}
\subsubsection{Video diffusion model.}
Several studies~\cite{blattmann2023align, wu2023tune, khachatryan2023text2video} propose to fine-tune the pre-trained text-to-image diffusion model to the video generation task. Although they exhibit promising results with minimal computational complexity, they often encounter challenges of temporal consistency at high frame rates. To alleviate this, recent methodologies~\cite{ho2022imagen, yu2023video, hu2023lamd, wang2023leo} train the diffusion model from scratch using video datasets. While these video diffusion models exhibit better temporal consistency and video quality, they struggle with memory inefficiency and controllability of video generation. To overcome these challenges, several methods focus on a low-dimensional latent space~\cite{yu2023video}, and an autoencoder module~\cite{hu2023lamd} to minimize computational complexity, while utilizing flow maps as a motion condition~\cite{wang2023leo} for conditional video generation. In this paper, we propose an efficient conditional video diffusion model, tailored for talking head generation.

%% file: assets/sec/3_Methodology.tex
\section{Preliminary- Denoising Diffusion Model}
Diffusion models~\cite{ho2020denoising, song2020denoising} approximate the data distribution by reconstructing the data sample from pure Gaussian noise through a gradual denoising process. Latent diffusion models~\cite{rombach2022high} perform this in latent space, not in RGB image space. Specifically, the RGB image $x_0$ is projected to the latent space $z_0$ using a pre-trained encoder, and $z_0$ is decoded back to $x_0$ using a pre-trained decoder. In the forward diffusion process, $z_0$ is gradually noisified into $z_t$ at time step $t \in \{1, \ldots,T\}$. The forward diffusion process is formulated with pre-defined variance $\beta_t$ such that:
\begin{equation}
z_t = \sqrt{\alpha_t}z_0 + \sqrt{1-\alpha_t}\epsilon, \quad \epsilon \sim \mathcal{N}(0,I),
\end{equation}
where $\alpha_t = \prod_{i=1}^{t} (1 - \beta_i)$.

The neural network $\mathcal{F}_\theta(z_t, t)$ is trained to reconstruct $z_0$ at any time step $t$. The objective function is defined by the Mean Squared Error (MSE) as follows:
\begin{equation}
    \label{training}
    \mathbb{E}_{z, t, \epsilon \sim \mathcal{N}(0,1)} \left[ \| z_0 - \mathcal{F}_\theta(z_t, t) \|^2_2 \right].
\end{equation}

After training, we can sample new data from the learned data distribution through an iterative denoising procedure, called the reverse diffusion process. In the reverse diffusion process, the neural network $\mathcal{F}_\theta(z_t, t)$ predicts the denoised latent $\hat{z}_{0,t}$ and converts it to $\hat{z}_{t-1}$ by the reparameterization trick~\cite{kingma2013auto}. This can be formulated as follows:
\begin{equation}
    \hat{z}_{t-1} = \sqrt{\alpha_{t-1}}\mathcal{F}_{\theta}(z_t, t) + \frac{\sqrt{1-\alpha_{t-1}-\sigma_t^2}}{\sqrt{1-\alpha_t}}(z_t - \sqrt{\alpha_{t}}\mathcal{F}_{\theta}(z_t, t)) + \sigma_t \epsilon,
\end{equation} 
where $\sigma_t$ at time step $t$ is the covariance of the Gaussian distribution.

By iteratively sampling $\mathcal{F}_\theta(z_t, t)$ starting from Gaussian noise $z_T$ during time step $t \in \{T, \ldots,1\}$, we obtain the new data sample $\hat{z}_0$ from the desired distribution. $\hat{z}_0$ is then decoded to an RGB image $\hat{x}_0$ by the pre-trained decoder.

\section{Method}

\input{assets/figures/fig_overall}

\subsection{Overview}
Our objective is to generate a talking head video $X = \{x^1, \ldots, x^n\}$ with lip movements synchronized with given audio $A = \{a^1, \ldots, a^n\}$ and the desired identity from an identity frame $x_{\mathrm{id}}$, where $n$ is the number of frames. To achieve this, MoDiTalker comprises two distinct diffusion models: Audio-to-Motion (AToM) and Motion-to-Video (MToV). AToM generates facial motion sequences, specifically facial landmarks $L = \{l^1, \ldots, l^n\}$ from input audio $A$, reflecting the speaker-specific facial structure from $x_{\mathrm{id}}$. Subsequently, MToV produces a realistic talking head video $X$, aligned facial motion sequences $L$ generated by AToM. Following prior works~\cite{shen2023difftalk}, MToV also uses the upper half of $X$ as pose frames $X_P = \{x^1_{P}, \ldots, x^n_{P}\}$ and focuses on generating lip movements in sync with the audio. For long video generation and providing identity cues, MToV leverages previously generated video clips as identity frames $X_I = \{x^1_{I}, \ldots, x^n_{I}\}$. The overall architecture of MoDiTalker is illustrated in Fig.~\ref{fig:inference}. 

\subsection{Audio-to-Motion (AToM) Diffusion Model}
Given audio input $A$ and a single identity frame $x_{\mathrm{id}}$, AToM aims to generate facial landmark sequences $L$ that are aligned with both the audio $A$ and the identity $x_{\mathrm{id}}$. For long motion generation, we use the first frame of the reference video as the identity frame $x_{\mathrm{id}}$. This allows us to generate subsequent motion clips by using the last motion frame of the previously generated clip as the initial facial landmark $l_{\mathrm{id}}$.
As illustrated in Fig.~\ref{fig:audio-to-landmark} (a), we introduce a modified transformer-based diffusion model based on~\cite{tseng2023edge}, with the audio $A$ and the identity frame $x_{\mathrm{id}}$ as conditions.

\input{assets/figures/fig_ATOM}

\vspace{-6pt}
\subsubsection{Architectural details.}
First, we design AToM to learn the residuals $\Delta L = \{\Delta l^1, \ldots, \Delta l^n\}$ from the initial facial landmark $l_{\mathrm{id}}$ to enhance stability and initialization. Specifically, we extract the initial facial landmark $l_{\mathrm{id}}$ from the identity frame $x_{\mathrm{id}}$ using a 3D Facial Morphable Model (3DMM)~\cite{blanz2023morphable}. The $l_{\mathrm{id}}$ is then passed through a trainable landmark encoder to produce the landmark embedding $F_L$. Note that in the training phase, we frontalize~\cite{ye2023geneface} the facial landmarks to ensure the model focuses on lip movements rather than pose.

We also extract the audio embedding $F_A = \{f^1, \ldots, f^n\}$ from the audio $A$, using HuBERT~\cite{hsu2021hubert} and a subsequent trainable audio encoder. Both landmark encoder and audio encoder are composed of simple transformer encoders. Both $F_L$ and $F_A$ are then integrated into the cross-attention modules of the diffusion model, by concatenating with time step embeddings. Benefiting from $l_{\mathrm{id}}$, which provides the speaker-specific facial structure, the model is enabled to capture speaker-agnostic representations (e.g., lip motions) from $F_A$. The landmark encoder and the audio encoder are both constructed using simple transformer encoders sharing the same architecture.

Moreover, to improve lip synchronization quality, we design AToM to separately process lip-related and lip-unrelated facial landmarks, as illustrated in Fig.~\ref{fig:audio-to-landmark}(b). We divide the facial landmarks into upper-half (lip-related) and lower-half (lip-unrelated) segments, then design the network to process them separately and subsequently merge them. The audio embedding $F_A$ is exclusively injected into the cross-attention module of the lip-related block, while the landmark embedding $F_L$ is conditioned after the two blocks are merged. This architectural choice enables the model to focus solely on lip movements, preserving the other facial regions, and ultimately enhancing lip synchronization fidelity. We provide an ablation study on different configurations in Tab.~\ref{tab:mcvdm_ablation} (a).

Finally, under the conditions of $F_L$ and $F_A$, the Gaussian noise $\Delta L_T$ is iteratively denoised through AToM to generate the denoised residual landmark sequences $\Delta \hat{L}_0$. The final facial motion sequences $L$ are obtained by adding $\Delta \hat{L}_0$ to the initial landmark $l_{\mathrm{id}}$.

\vspace{-6pt}
\subsubsection{Training.} In training, the diffusion model is trained to reconstruct $\Delta L$, using the initial landmark embedding $F_L$ to provide speaker-specific facial structures and the audio embedding $F_A$ for lip movements, as conditions. Therefore, we redefine Equ.~\ref{training} for the objective function of training AToM, $\mathcal{F}_\mathrm{AToM}$, as follows:
\begin{equation}
    \label{training_landmarks}
    \mathcal{L}_{\mathrm{AToM}} = \mathbb{E}_{\Delta L, t, \epsilon \sim \mathcal{N}(0,1), F_L, F_A} \left[ \| \Delta L_0 - \mathcal{F}_{\mathrm{AToM}}(\Delta L_t, t; F_L, F_A) \|^2_2 \right].
\end{equation}
Here, $\Delta L_0$ is defined as the difference between facial landmarks from ground-truth video frames and those from an identity frame, and $\Delta L_t$ is the noisified version of $\Delta L_0$ at time step $t$. Additionally, $F_L$ is derived from an identity frame $x_{\mathrm{id}}$, which corresponds to the first frame of the ground-truth videos.

\subsection{Motion-to-Video (MToV) Diffusion Model}
MToV aims to generate realistic talking head videos $X$, mirroring desired identities and audio inputs. To achieve this, we use three distinct conditions: facial landmark sequences $L$, pose frames $X_P$, and identity frames $X_I$. Previous diffusion-based methods~\cite{stypulkowski2024diffused, shen2023difftalk} generate videos in a frame-by-frame manner, resulting in sub-optimal temporal consistency and identity preservation due to the stochastic nature of the diffusion model. 

To tackle this, as illustrated in Fig.~\ref{fig:inference}, we introduce a modified video diffusion model based on~\cite{yu2023video} with given conditions. Conditioning the video diffusion model traditionally requires 4-dimensional representations~\cite{ho2022video}, which pose significant computational challenges during training and inference. To overcome this limitation, we adopt tri-plane representations~\cite{yu2023video, ho2022video} to effectively employ a video diffusion model tailored for talking face generation.

\vspace{-6pt}
\subsubsection{Architectural details.}
First, AToM synchronizes facial landmark sequences $L \in \mathbb{R}^{S \times C \times H \times W}$ with audio and identity, where $S$, $C$, $H$, and $W$ represent the number of sequences, channels, height, and width, respectively. These sequences are then encoded into tri-plane representations $Z_L = \{ z_l^{hw}, z_l^{hs}, z_l^{ws} \}$ by the landmark encoder $\encoder_L$, where $z_l^{hw} \in \mathbb{R}^{c \times h \times w}$, $z_l^{hs} \in \mathbb{R}^{c \times h \times s}$, and $z_l^{ws} \in \mathbb{R}^{c \times w \times s}$. Here, $c$, $h$ and $w$ denote the embedded channel, height, and width, respectively. Specifically, $z_l^{hw}$ provides the model with speaker-specific facial structures, while $z_l^{hs}$ and $z_l^{ws}$ encode the temporal relationships between frames. Further detailed visualizations related to tri-plane representations are provided in the Appendix Sec.~\textcolor[rgb]{1.0,0,0}{B.1} and Sec.~\textcolor[rgb]{1.0,0,0}{B.2}.

Simultaneously, following previous works~\cite{prajwal2020lip, shen2023difftalk}, we condition the model on pose frames $X_P \in \mathbb{R}^{S \times C \times H \times W}$, corresponding to the upper part of the desired video $X$. We further leverage head poses from the pose frames $X_P$ to achieve facial landmarks alignment. We achieve this by applying an affine transformation~\cite{wang2021one} to the frontalized facial landmarks from AToM, aligning them with the desired head pose. Detailed process of transformations is explained in the Appendix Sec.~\textcolor[rgb]{1.0,0,0}{A.2}. $X_P$ is then encoded into tri-plane representations $Z_P = \{ z_p^{hw}, z_p^{hs}, z_p^{ws} \}$ through the pose encoder $\encoder_P$. This allows the model to focus solely on generating lip movements while preserving other facial regions, ultimately improving lip synchronization.

Furthermore, to ensure temporal consistency in extended video generation and provide identity cues, we utilize previous video clips as identity frames $X_I$, following~\cite{prajwal2020lip}. These frames are processed through the identity encoder $\encoder_I$, resulting in tri-plane representations $Z_I = \{ z_i^{hw}, z_i^{hs}, z_i^{ws} \}$. 

Finally, we concatenate $Z_L$, $Z_P$, and $Z_I$ along the channel axis. Given the conditions of $Z_L$, $Z_P$, and $Z_I$, the pure Gaussian noise $z_T$ is gradually denoised through the MToV process, generating the denoised video latent $\hat{z}_0$. This is then decoded back into the RGB space, $\hat{X}_0$, by a pre-trained decoder.

By combining an efficient architectural design with carefully selected conditions, MToV can generate high-fidelity talking head videos with enhanced temporal consistency and identity preservation, all while requiring substantially lower computational complexity. Our design choices are further detailed in Tab.~\ref{tab:mcvdm_ablation}.

\vspace{-6pt}
\subsubsection{Training encoders.} MToV includes three distinct encoders: the landmark encoder $\encoder_L$, identity encoder $\encoder_I$, and pose encoder $\encoder_P$. All encoders compress cubic-like 4D video inputs $V \in \mathbb{R}^{S \times C \times H \times W}$ into image-like three 2D latents $\in \mathbb{R}^{c \times h \times w}$, where $S$, $C$, $H$, and $W$ represent sequence, channel, height, and width, while $c$, $h$, and $w$ represent embedded channel, height, and width, respectively. For simplicity, we denote $\encoder$ to include all three encoders $\encoder_L$, $\encoder_I$, and $\encoder_P$.

The encoder $\encoder$ is trained using an autoencoder~\cite{bertasius2021space}, where $\encoder$ embeds $V$ in the RGB space into the latent space, and the corresponding decoder $\decoder$ reconstructs the latent space back into $\hat{V}$ in the RGB space.

In the training phase, we use two different losses: pixel-level reconstruction loss~\cite{zhao2016loss} and the perceptual loss~\cite{zhang2018unreasonable}, which enforce the encoder $\encoder$ and decoder $\decoder$ to accurately embed and reconstruct the given inputs in both image space and feature space. The total objective function is formulated as:
\begin{equation}
\mathcal{L}_{\text{encoder}} = \lambda_{1}\mathbb{E}_{V}\left[\|V-\hat{V}\|_{1}\right] + \lambda_{2} \mathbb{E}_{V}\left[\|\phi(V)-\phi(\hat{V})\|_{1}\right],
\end{equation}
where $\phi$ denotes the perceptual feature extractor~\cite{zhang2018unreasonable}. For all our experiments, we carefully select $\lambda_{1} = \lambda_{2} = 1$.

\vspace{-6pt}
\subsubsection{Training diffusion model.}
In training, the video diffusion model learns to reconstruct talking head videos $X$, using landmark embeddings $Z_L$, pose embeddings $Z_P$, and identity embeddings $Z_I$ as conditions. We can redefine Equ.~\ref{training} for training MToV, $\mathcal{F}_{\mathrm{MToV}}$, as follows:
\begin{equation}
    \label{training_videos}
    \mathcal{L}_{\mathrm{MToV}} =\mathbb{E}_{X, t, \epsilon \sim \mathcal{N}(0,1), Z_L, Z_I, Z_P} \left[ \| X_0 - \mathcal{F}_{\mathrm{MToV}}(X_t, t; Z_L, Z_I, Z_P) \|^2_2 \right],
\end{equation}
where $X_0$ is the ground-truth talking head video, and $X_t$ is the noised $X_0$ at time step $t$. 

%% file: assets/figures/fig_overall.tex
\begin{figure*}[t!]
  \centering
  \includegraphics[width=1.00\linewidth]{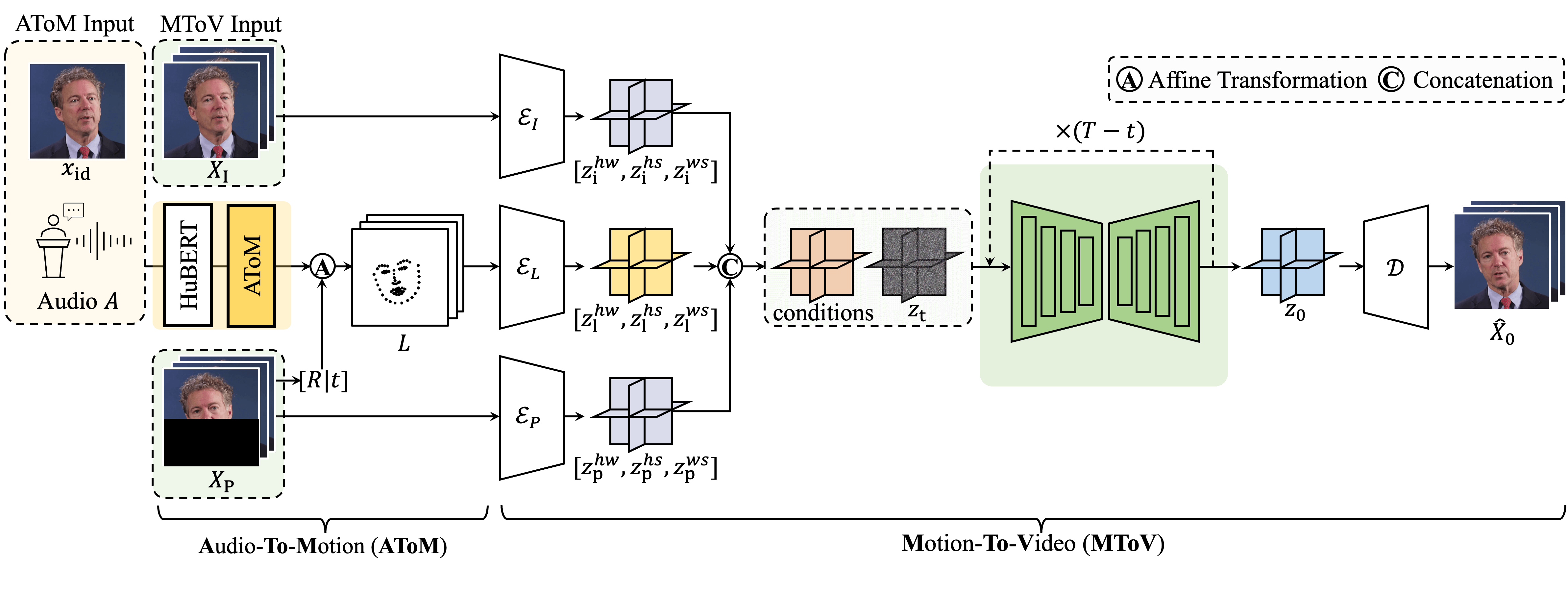}
  \vspace{-20pt}
  \caption{\textbf{Overall network architecture of MoDiTalker.} Our framework consists of two distinct diffusion models: Audio-to-Motion (AToM) and Motion-to-Video (MToV). AToM aims to generate lip-synchronized facial landmarks, given an identity frame $x_{\mathrm{id}}$ and audio input $A$, as conditions. MToV generates high-fidelity talking head videos $\hat{X}_0$ using synthesized facial landmarks $L$ from AToM, identity frames $X_\mathrm{I}$, and pose frames $X_\mathrm{P}$ as conditions.}
  
\label{fig:inference}
\vspace{-5pt}
\end{figure*}

%% file: assets/figures/fig_ATOM.tex
\begin{figure*}[t]
    \centering
    \includegraphics[width=1.00\linewidth]{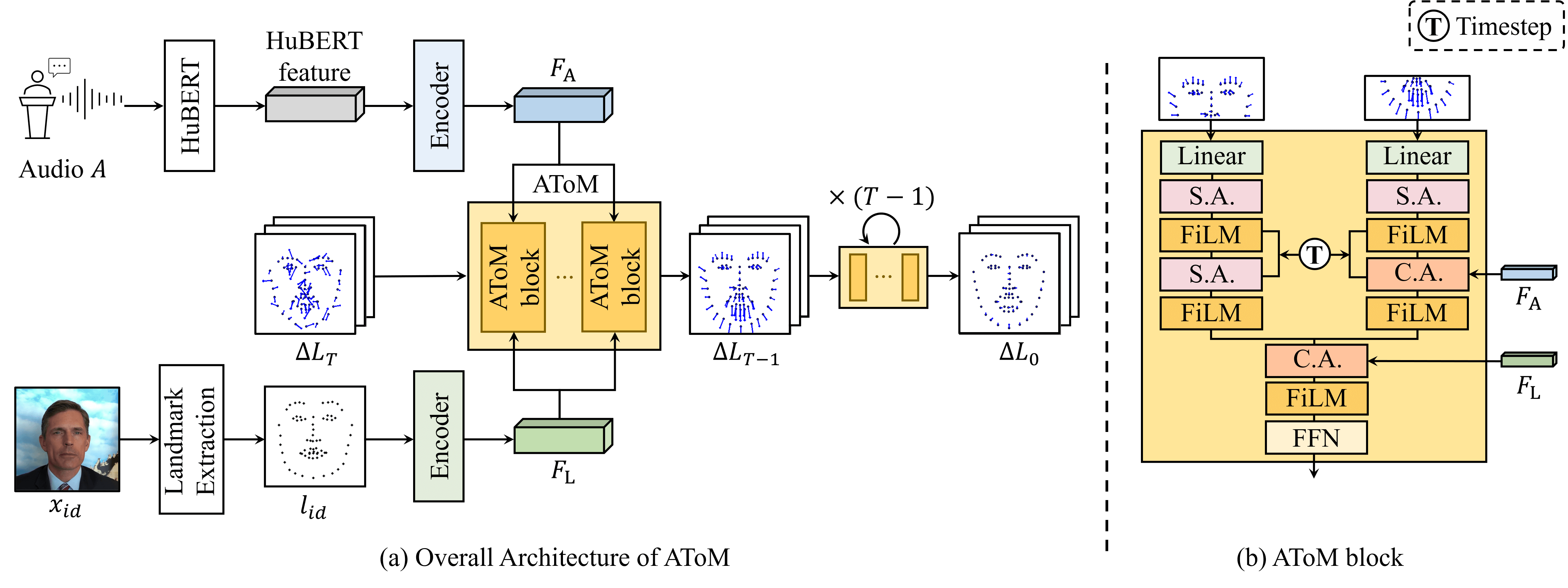}
    \vspace{-20pt}
    \caption{\textbf{Overview of the Audio-to-Motion (AToM) diffusion model:} (a) AToM is a transformer-based diffusion model that learns the residual between the initial landmark $l_\mathrm{init}$ and the landmark sequence, using the audio embedding $F_A$ and the initial landmark embedding $F_L$ as conditions. In addition, (b) we design AToM block to process lip-related (upper-half) and lip-unrelated (lower-half) landmarks separately, allowing the model to focus more on generating lip-related movements while preserving the facial shape of the speaker.}

\label{fig:audio-to-landmark}
\vspace{-5pt}
\end{figure*}

%% file: assets/sec/4_Experiments.tex
\section{Experiments}
\subsection{Experimental Settings}
\subsubsection{Implementation details.}
For all experiments, we used 1 NVIDIA RTX 3090 GPU. For AToM, we train the model for 300k iterations with a learning rate of 1e-4. For MToV, we train the model for 600k iterations with a learning rate of 1e-4. To alleviate jittering, we employed a blending technique using Gaussian blur, as described in ~\cite{chen2020simswap}. Additional implementation details are provided in the Appendix \textcolor[rgb]{1.0,0,0}{A.1} and \textcolor[rgb]{1.0,0,0}{A.2}.

\vspace{-6pt}
\subsubsection{Datasets.}
We used the LRS3-TED~\cite{afouras2018lrs3} and HDTF~\cite{zhang2021flow} datasets to train our AToM and MToV models, respectively. The LRS3-TED dataset comprises 400 hours of TED talk videos, providing a large lip-reading corpus. For AToM, we extracted video frames at 25 fps and audio at a 16,000 Hz sampling rate. For MToV, we randomly selected 312 videos from the HDTF dataset for training, using the remaining 98 videos for testing.

\input{assets/tables/main_comparison}
\input{assets/figures/qual_main}

\input{assets/figures/qual_cross_id}

\vspace{-6pt}
\subsubsection{Evaluation metrics.}
We evaluated our methods using several metrics predominantly used in this field~\cite{Seitzer2020FID, heusel2017gans, narvekar2011no, zhang2018unreasonable, chung2017out, chen2018lip}. In detail, we used \textbf{FID} ($\downarrow$)~\cite{Seitzer2020FID} and \textbf{PSNR} ($\uparrow$)~\cite{heusel2017gans} to assess the image fidelity. Also, we used \textbf{CPBD} ($\uparrow$)~\cite{narvekar2011no} to evaluate the sharpness of the generated frames, \textbf{LPIPS} ($\downarrow$)~\cite{zhang2018unreasonable} to measure the visual resemblance, and \textbf{CSIM}(↑) to examine the identity preservation. For lip-sync quality, we utilized SyncNet~\cite{chung2017out} scores: \textbf{LSE-D} ($\downarrow$)~\cite{chung2017out}, measuring the distance between lip and audio features, and \textbf{LSE-C}($\uparrow$)~\cite{chung2017out}, measuring the confidence score between them. Finally, we used \textbf{LMD} ($\downarrow$)~\cite{chen2018lip} which measures the accuracy of generating lip movements.

\subsection{Qualitative Results}
\input{assets/figures/qual_difftalk}

In Fig.~\ref{fig:main_comparison_qual}, we present a qualitative comparison with prior GAN-based~\cite{prajwal2020lip, zhou2021pose, zhou2020makelttalk, wang2021audio2head} and diffusion-based~\cite{stypulkowski2024diffused, ma2023dreamtalk} studies, highlighting the superiority of MoDiTalker. We observe that the previous GAN-based approaches exhibit sub-optimal video quality. Specifically, Wav2Lip~\cite{prajwal2020lip} and Audio2Head~\cite{wang2021audio2head} produce blurry outputs, while MakeItTalk~\cite{zhou2020makelttalk} and PC-AVS~\cite{zhou2021pose} generate jittery videos. Meanwhile, DreamTalk~\cite{ma2023dreamtalk}, achieve reasonable video quality compared to GAN-based methods but suffer from issues with identity preservation and temporal consistency since they generate videos in a frame-by-frame manner. Diffused Heads~\cite{stypulkowski2024diffused}, trained only on the CREMA~\cite{cao2014crema}, performs poorly on the HDTF dataset being compromised by error accumulation across the generated frames. Unlike prior approaches, MoDiTalker generates high-quality videos while preserving identity and temporal consistency. Comparisons on the unseen dataset~\cite{cao2014crema} can be found in Appendix \textcolor[rgb]{1.0,0,0}{C.3}.

Moreover, in Fig.~\ref{fig:qual_cross_id}, we also present a qualitative comparison with diffusion-based studies~\cite{stypulkowski2024diffused, ma2023dreamtalk} under cross-identity scenarios, which use the input audio from a different identity than the identity frames. Prior diffusion-based methods commonly inject the audio embedding through a cross-attention module in diffusion U-Net. This prevents the model from separating speaker-specific and speaker-agnostic representations, such as facial shape and phoneme, leading to sub-optimal performance in cross-identity scenarios. In contrast, MoDiTalker shows superior generalization capability, benefiting from a distinct training process that extracts intermediate facial landmarks from given audio to generate the final talking video. Additionally, we provide a qualitative comparison with previous GAN-based studies~\cite{prajwal2020lip, zhou2021pose, zhou2020makelttalk, wang2021audio2head} in the Appendix \textcolor[rgb]{1.0,0,0}{C.1}.

\subsection{Quantitative Results}
\input{assets/figures/qual_genface}
In Tab.~\ref{tab:compare}, we present a quantitative comparison with existing talking head generation approaches~\cite{prajwal2020lip,zhou2021pose,zhou2020makelttalk,wang2021audio2head,stypulkowski2024diffused,ma2023dreamtalk} using the HDTF dataset~\cite{zhang2021flow}. MoDiTalker significantly outperforms current state-of-the-art methods across all video quality metrics and the LMD score by a large margin. Notably, our model also achieves competitive performance in the LSE-D score without utilizing SyncNet in our training, while previous works~\cite{prajwal2020lip,ma2023dreamtalk} incorporate SyncNet loss during their training phase. This demonstrates the superior generalizability of our method. Note that we do not include a comparison with DiffTalk~\cite{shen2023difftalk} in our evaluation. As noted in~\cite{shen2023difftalk}, despite our efforts, we were unable to reproduce the results reported in their paper.

This success is attributed to our framework design, AToM and MToV, with each component further discussed in Sec.~\ref{sec:ablation_study}. Furthermore, our audio-to-motion landmark generator, AToM, surpasses the state-of-the-art landmark generative model, GeneFace~\cite{ye2023geneface}, in lip synchronization and identity preservation. This is further analyzed in Sec.~\ref{sec:ablation_study}, and detailed in Tab.~\ref{tab:mcvdm_ablation}, Fig.~\ref{fig:atom_ablation_qual} and Fig.~\ref{fig:land_comparison}.

\input{assets/figures/qual_user_study}

\subsection{User Study}
We conducted a user study to compare MoDiTalker with previous GAN-based~\cite{prajwal2020lip, zhou2021pose, zhou2020makelttalk, wang2021audio2head} and diffusion-based~\cite{stypułkowski2023diffused, shen2023difftalk ,ma2023dreamtalk} works, focusing on lip-sync fidelity, identity preservation, and video quality. The results are summarized in Fig.~\ref{fig:user_study}. For lip-sync fidelity, participants were provided with audio input and generated talking head videos from different methods. For identity preservation, we provided an identity frame and generated talking head videos from various methods. Lastly, for video quality, we presented generated talking head videos from different studies. A total of 87 participants responded to 15 samples for each question, totaling 1,305 responses for samples. Note that we randomly selected 5-second segments of the generated data, using HDTF audio and identity frames. MoDiTalker surpasses previous GAN-based and diffusion-based works in all aspects including lip-sync fidelity, identity preservation, and video quality, further highlighting the superiority of our method.

\subsection{Ablation Study}
\label{sec:ablation_study}

\subsubsection{Component analysis of AToM diffusion model.} In this ablation study, we aim to show the effectiveness of each component of AToM. Fig.~\ref{fig:atom_ablation_qual} and Tab.~\ref{tab:mcvdm_ablation} (a) summarize the qualitative and quantitative results, respectively. We also compare AToM with GeneFace, the state-of-the-art audio-to-landmark generative model. The baseline \textbf{(I)} presents a diffusion model which generates facial landmark sequences $L$ using only the audio condition $F_A$. \textbf{(II)} represents the residual prediction, estimating residuals $\Delta{L}$ from initial landmark sequences $L_\mathrm{init}$. Compared to \textbf{(I)}, \textbf{(II)} significantly enhances Lip-sync scores, benefiting from enhanced initialization and training stability. \textbf{(III)} incorporates initial landmark embedding $F_L$ as a condition, which markedly boosts LMD scores and the accuracy of the facial shape in the desired identity $x_\mathrm{id}$. Lastly, the disentangled framework \textbf{(IV)}, detailed in the right part of Fig.~\ref{fig:audio-to-landmark}, separates lip-related and unrelated segments. With disentangled attention, AToM concentrates on lip movements while preserving other facial segments, leading to its highest performance. Combining these, in comparison with GeneFace, AToM achieves superior performance in LMD and Lip-sync scores, demonstrating its capability to generate high-quality facial landmarks given audio and identity frames. This is further evident in the qualitative comparison of AToM with GeneFace, as shown in Fig.~\ref{fig:land_comparison}.

\input{assets/tables/abl_mtov}
\input{assets/figures/qual_abl_atom}

\vspace{-7pt}
\subsubsection{Component analysis of MToV diffusion model.}
We also evaluate different configurations of MToV, as presented in Tab.~\ref{tab:mcvdm_ablation} (b) and Figure~\ref{fig:mcvdm_ablation_qual}. This ablation study highlights the significance of different conditions. The baseline model \textbf{(I)}, without any conditions, learns the general data distribution but fails to capture the desired lip synchronization, identity, and head pose. Introducing identity frames $X_\mathrm{I}$ as a condition in configuration \textbf{(II)} ensures the model preserves the same identity throughout the generated video. Moreover, incorporating pose frames $X_\mathrm{P}$ as additional conditions in \textbf{(III)} not only focuses the generation on the mouth region but also accurately captures the head pose, which notably improves both video quality (CSIM, CPBD) and lip-sync metrics (LMD). The introduction of audio embedding through cross-attention layers in configuration \textbf{(IV)} roughly aligns with the lip movements but may lack precision. In contrast, utilizing facial motion landmarks $L$ from AToM \textbf{(V)} significantly enhances lip-sync accuracy, demonstrating the effectiveness of leveraging intermediate facial motion to generate high-fidelity talking head videos. By utilizing these appropriate conditions, MToV has the ability to generate high-quality videos with accurate lip synchronization.

\input{assets/figures/qual_abl_mtov}

\vspace{-5pt}
\subsubsection{Computational complexity.} We conducted an efficiency comparison with prior diffusion-based models. Although several studies~\cite{shen2023difftalk, stypułkowski2023diffused} have successfully employed diffusion models to generate talking head videos, their frame-by-frame generation approach poses a significant time-inefficiency. For instance, DiffTalk~\cite{shen2023difftalk} and Diffused Heads~\cite{stypułkowski2023diffused} require 1003 and 716 seconds, respectively, to produce a 5-second video at 25 fps on a single Nvidia 24GB 3090 RTX GPU. In contrast, our model needs only 23 seconds, demonstrating a significant efficiency advantage. Our method is 43 times faster than DiffTalk and 31 times faster than Diffused Heads. This speed gain is especially valuable for producing longer videos, making our approach much more practical for real-world use.

%% file: assets/tables/main_comparison.tex
\begin{table*}[t]
\setlength{\tabcolsep}{2pt}
\renewcommand{\arraystretch}{1.2}
\centering
\vspace{-10pt}
\caption{\textbf{Quantitative comparison with previous works on video quality and lip synchronization:} We compare MoDiTalker with previous GAN-based methods, including Wav2Lip~\cite{prajwal2020lip}, PC-AVS~\cite{zhou2021pose}, MakeItTalk~\cite{zhou2020makelttalk}, and Audio2Head~\cite{wang2021audio2head}, as well as diffusion-based methods, including DiffusedHead~\cite{stypułkowski2023diffused} and DreamTalk~\cite{ma2023dreamtalk} on the HDTF~\cite{zhang2021flow} test dataset.}
\resizebox{0.85\linewidth}{!}
{\begin{tabular}{l c c c c c c c}
\toprule

& \multicolumn{5}{c}{Video Quality} 
& \multicolumn{2}{c}{Lip Sync.} \\
\cmidrule(lr){2-6} \cmidrule(lr){7-8}
 & {FID $\downarrow$} & {CPBD $\uparrow$} & {PSNR $\uparrow$} & {LPIPS $\downarrow$} & {CSIM $\uparrow$}
& {LMD $\downarrow$} & {LSE-D $\downarrow$} \\ 
\midrule\midrule
Real Video 
& {0.00} &{0.43} & {-} &{-} &{1.00}  &{0.00} &{6.98} \\ \midrule
Wav2Lip~\cite{prajwal2020lip} 
& {16.34} & {0.35} &{34.81} & {0.03} &{0.90} &{1.43}&\textbf{5.86} \\
PC-AVS~\cite{zhou2021pose} 
& {117.85} &{0.29} &{28.23}& {0.38} &{0.35} &{9.36} &{7.06}\\
MakeItTalk~\cite{zhou2020makelttalk} 
&{66.21} &{0.43} &{29.87} & {0.16} &{0.82} &{3.46} &{10.23} \\
Audio2Head~\cite{wang2021audio2head} 
& {65.96} &{0.37} &{29.85} & {0.19} &{0.71} &{4.33} &{7.49} \\ \midrule
DiffusedHead~\cite{stypułkowski2023diffused}
& {192.74} &{0.11} &{28.21} & {0.274} &{0.16} &{22.19} &{12.37} \\ 
DreamTalk~\cite{ma2023dreamtalk}
& {124.51} &{0.43} &{29.59} & {0.20} &{0.72} &{2.56} &{8.37} \\ \hlrow
\textbf{MoDiTalker} 
& \textbf{14.15} & \textbf{0.46} & \textbf{35.82}  &\textbf{0.01} & \textbf{0.92} &\textbf{1.38}  &{9.15} \\
\bottomrule
\end{tabular}}
\vspace{-5pt}
\label{tab:compare}
\end{table*}

%% file: assets/figures/qual_main.tex
\begin{figure}[t]
    \centering
    \includegraphics[width=1.0\linewidth]{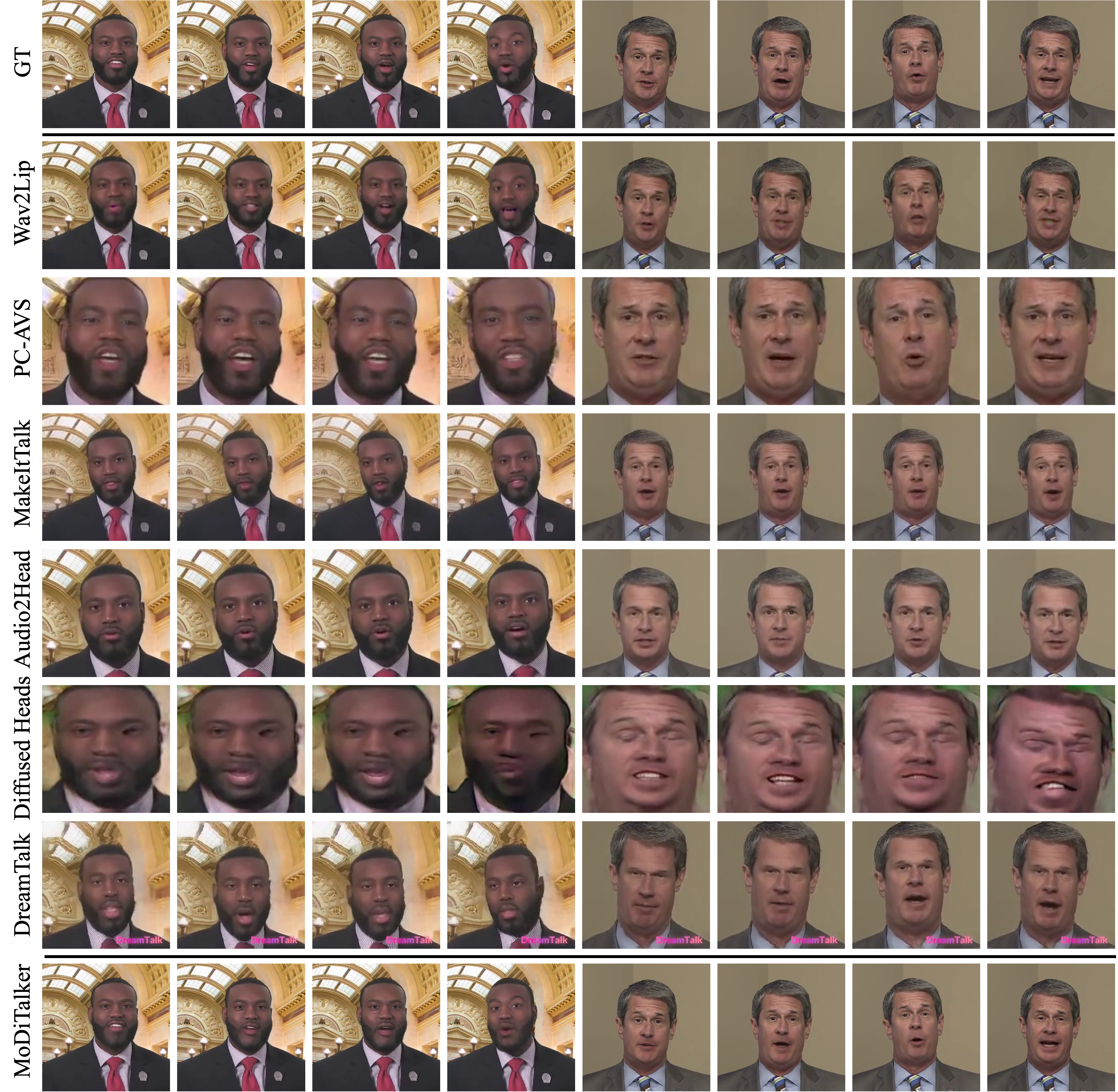}
    
    \vspace{-5pt}
    \caption{\textbf{Qualitative comparison with previous works:} We compare MoDiTalker with previous GAN-based methods, including Wav2Lip~\cite{prajwal2020lip}, PC-AVS~\cite{zhou2021pose}, MakeItTalk~\cite{zhou2020makelttalk}, and Audio2Head~\cite{wang2021audio2head}, as well as diffusion-based methods, including Diffused Heads ~\cite{stypulkowski2024diffused} and DreamTalk~\cite{ma2023dreamtalk}.}
    \vspace{-15pt}
\label{fig:main_comparison_qual}
\end{figure}

%% file: assets/figures/qual_cross_id.tex
\begin{figure}[t]
    \centering
    \includegraphics[width=1.0\linewidth]{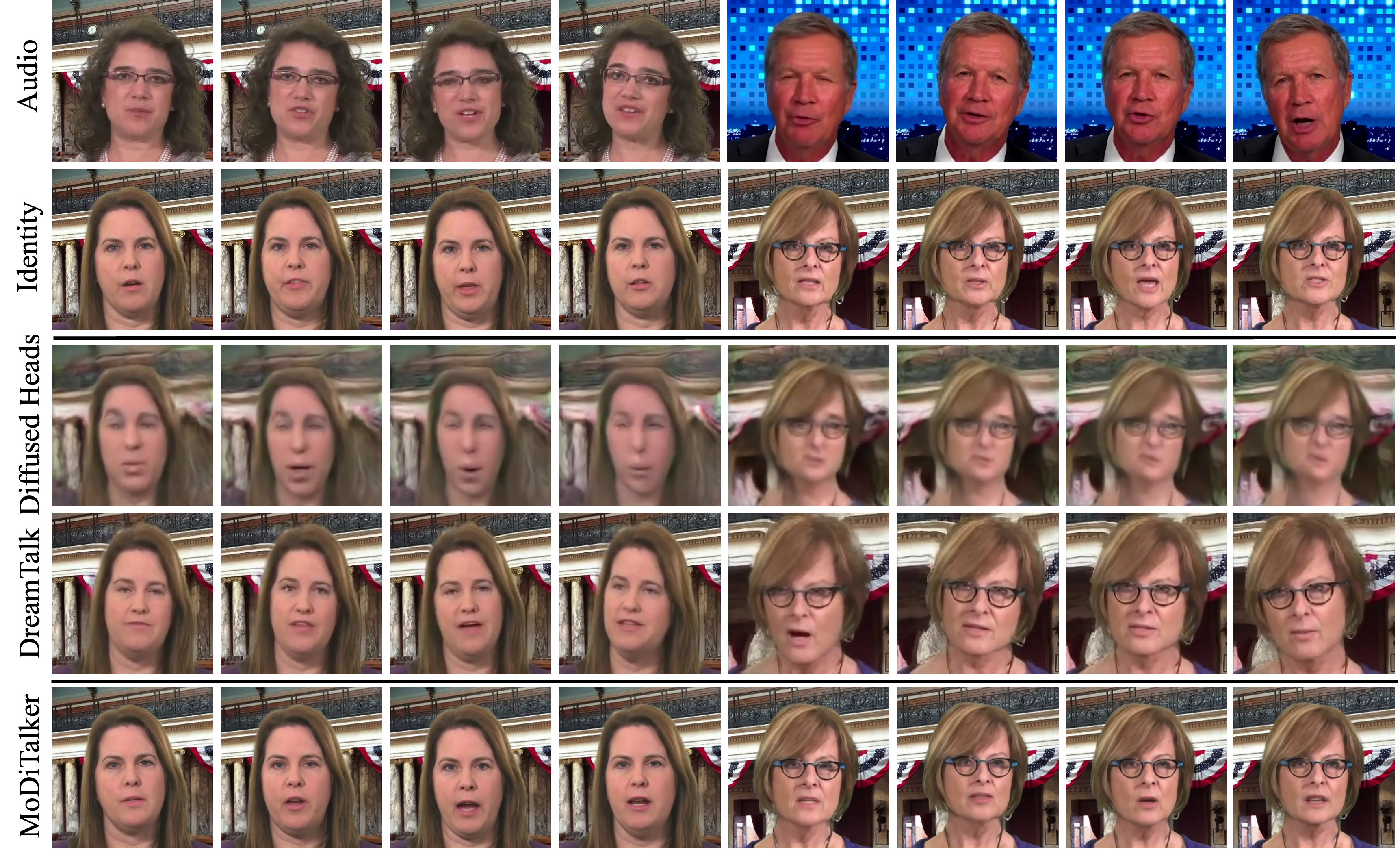}
    
    \vspace{-5pt}
    \caption{\textbf{Qualitative comparison with previous works on cross identity setting:} We compare MoDiTalker with previous  diffusion-based methods, including Diffused Heads\cite{stypulkowski2024diffused} and DreamTalk\cite{ma2023dreamtalk}.}

\label{fig:qual_cross_id}
\vspace{-10pt}
\end{figure}

%% file: assets/figures/qual_difftalk.tex
\begin{wrapfigure}{r}{6cm}
    \vspace{-20pt}
    \centering
    \includegraphics[width=1\linewidth]{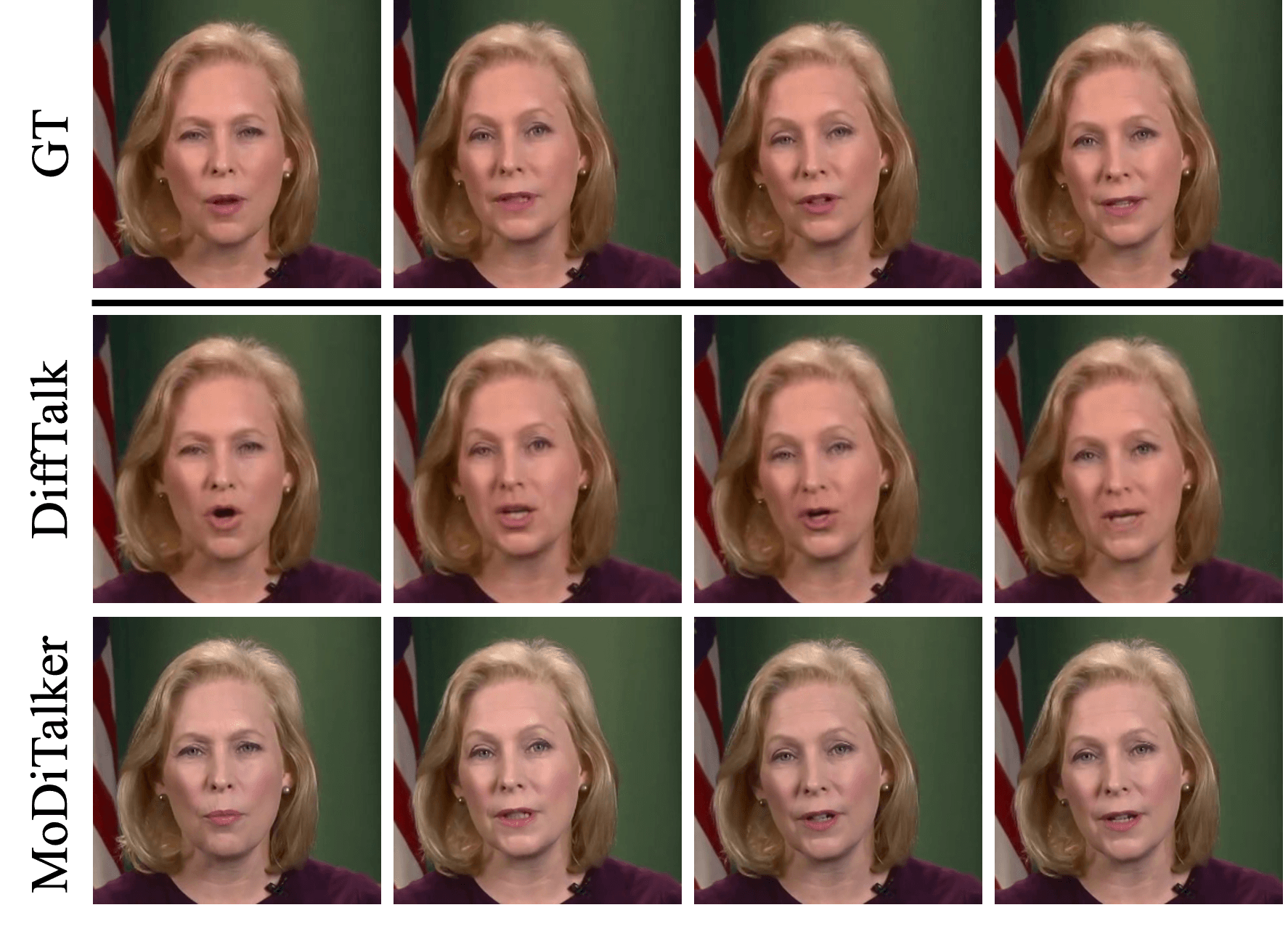}
    \vspace{-20pt}
    \caption{\textbf{Qualitative comparison with DiffTalk~\cite{shen2023difftalk}.}}
\label{fig:main_comparison_difftalk}
\vspace{-15pt}
\end{wrapfigure}

%% file: assets/figures/qual_genface.tex
\begin{wrapfigure}{r}{6cm}
    \centering
    \vspace{-20pt}
    \includegraphics[width=1\linewidth]{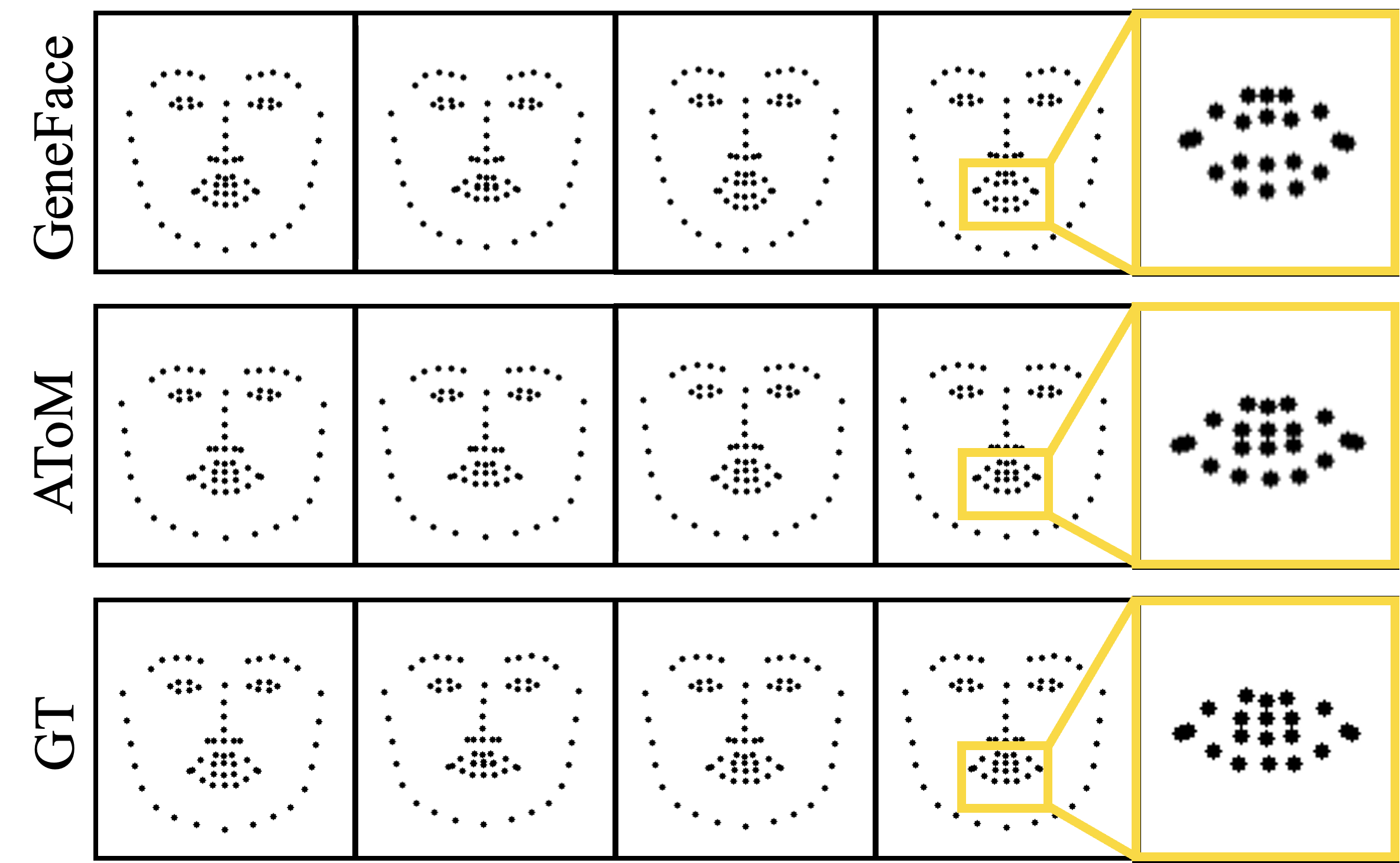}
    \caption{\textbf{Qualitative comparison with GeneFace~\cite{ye2023geneface}.}}
\label{fig:land_comparison}
\vspace{-19pt}
\end{wrapfigure}

%% file: assets/figures/qual_user_study.tex
\begin{figure}[t]
\centering
\includegraphics[width=1\linewidth]{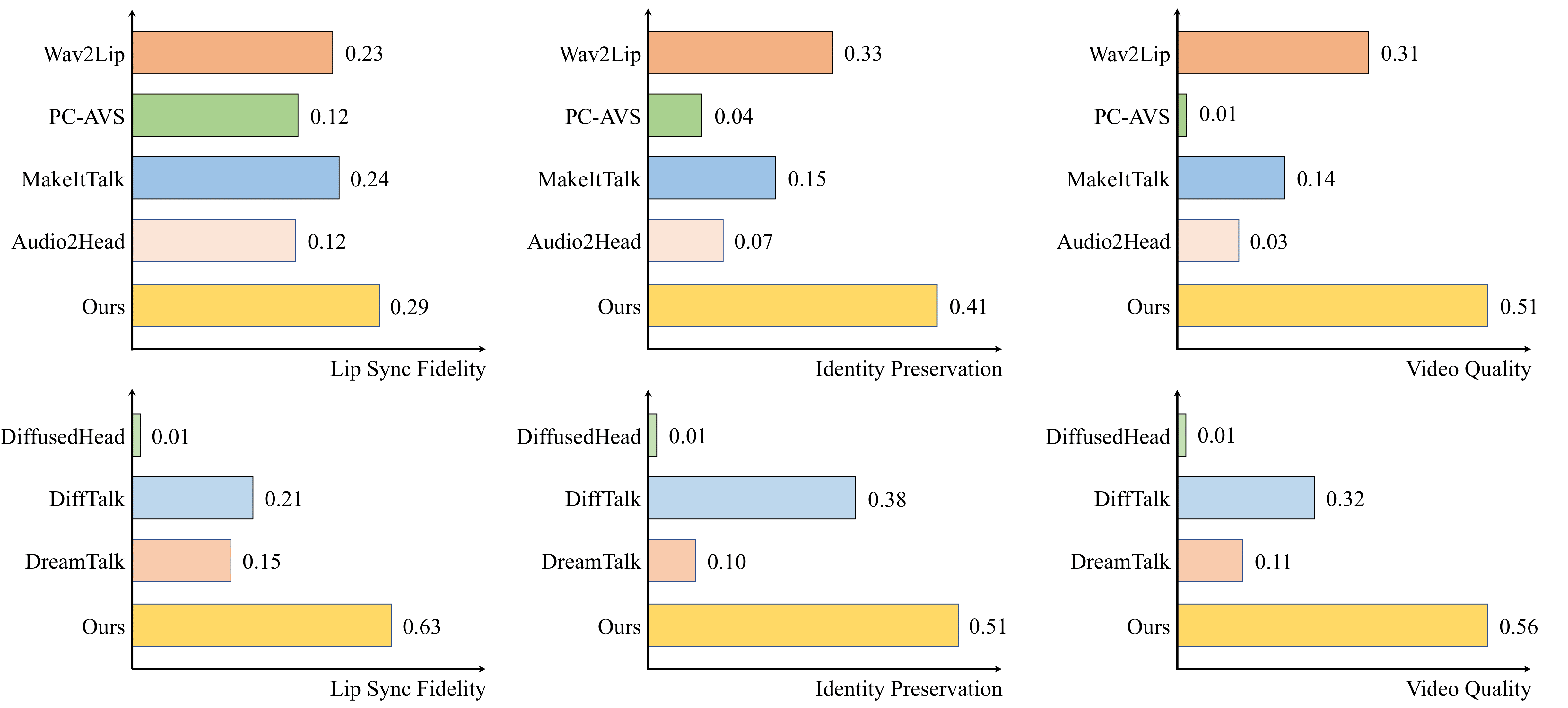}
   \caption{\textbf{User study results.} We compare MoDiTalker with GAN-based models~\cite{prajwal2020lip, zhou2021pose, zhou2020makelttalk, wang2021audio2head} (top) and diffusion-based models~\cite{stypułkowski2023diffused, shen2023difftalk ,ma2023dreamtalk} (bottom).}
	\label{fig:user_study}
\end{figure}

%% file: assets/tables/abl_mtov.tex
        
        
    
        


\begin{table*}[t]
  \centering
  \caption{\textbf{Component analysis on AtoM and MToV.}}
  \vspace{-15pt}
  \resizebox{1\linewidth}{!}{
  \begin{subtable}{.5\textwidth}
  \raggedleft
  \resizebox{1\linewidth}{!}{
    \begin{tabular}[t]{l l ccccc}\toprule
    
        & Component & LMD $\downarrow$  & LSE-C $\uparrow$   \\\midrule 
        
        & GeneFace~\cite{ye2023geneface} &{1.41} & {0.339}\\ \midrule
        
        \textbf{(I)} & Baseline  & {1.63} & {0.316}\\ 
        
        \textbf{(II)} & 
        (I) + Residual. &{1.47} &{0.386} \\
        
        \textbf{(III)} & 
        (II) + $F_L$ & {1.31} & {0.385}\\ \hlrow
        
        \textbf{(IV)} & 
        (III) + Disen. Att. (\textbf{Ours}) &\textbf{1.26} &\textbf{0.390} 
        
        \\ \bottomrule
    \end{tabular}}
    \caption{\textbf{Audio-to-Motion model (AToM)} }
  \end{subtable}
  
  \begin{subtable}{.525\textwidth}
    \raggedright
    \resizebox{1\linewidth}{!}{
    \begin{tabular}[t]{llcccc}\toprule
        & Component &  CSIM $\uparrow$ & CPBD $\uparrow$  & LMD $\downarrow$    \\\midrule 
        
        \textbf{(I)} 
        & Baseline &{-} & {-} & {-}\\ 
        
        \textbf{(II)} &
        (I) + $X_I$ &{0.87}  &{0.42} & {11.18} \\
    
        \textbf{(III)} &
        (II) + $X_P$
        &{0.91} & {0.44}  & {5.23} \\ 
        
        \textbf{(IV)} &
        (III) + A  &{0.89}  & {0.43}  & {2.67} \\

        \hlrow
        \textbf{(V)} &
        (III) + $L$ (\textbf{Ours})  &\textbf{0.92}  & \textbf{0.46} &\textbf{1.38}  \\
        \bottomrule
    \end{tabular}} 
    \caption{\textbf{Motion-To-Video model (MToV)}}
  \end{subtable}}
  \vspace{-10pt}
  \label{tab:mcvdm_ablation}
\end{table*}

%% file: assets/figures/qual_abl_atom.tex
\begin{figure}[t]
    \centering
\includegraphics[width=1.0\linewidth]{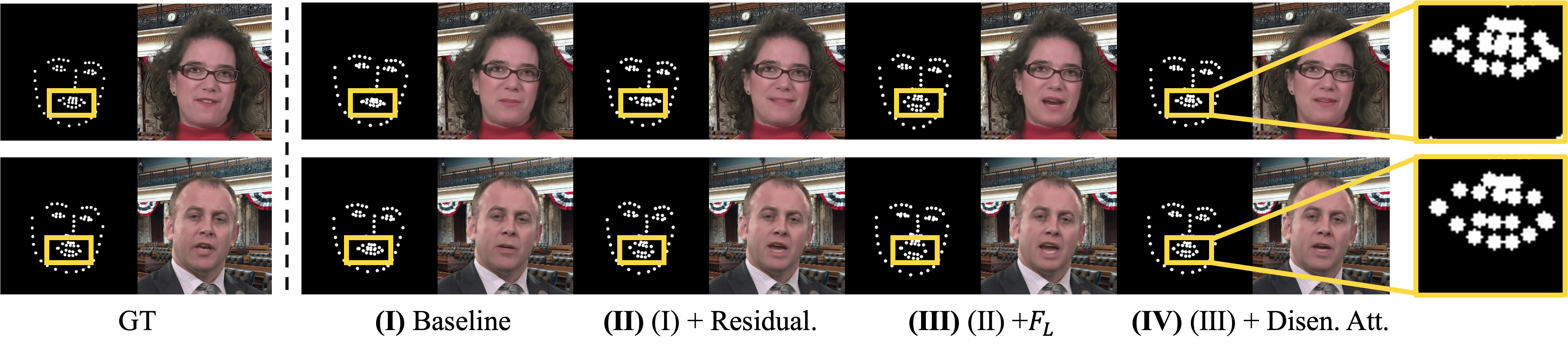}
    \vspace{-20pt}
    \caption{\textbf{Ablation study for AToM.} Baseline (I) generates facial landmark sequences $L$ using only audio condition $F_\mathrm{A}$. Predicting residuals $\Delta{L}$ (II) significantly improves lip sync capacity. In (III), using the initial landmark embedding $F_\mathrm{L}$ as a condition enhances identity preservation with precise facial contours. In (IV), disentangled attention framework separating lip-related and unrelated segments leads to optimal outcomes.}
\label{fig:atom_ablation_qual}
\vspace{-15pt}
\end{figure}

%% file: assets/figures/qual_abl_mtov.tex
\begin{figure*}[t]
    \centering
    \includegraphics[width=1.0\linewidth]{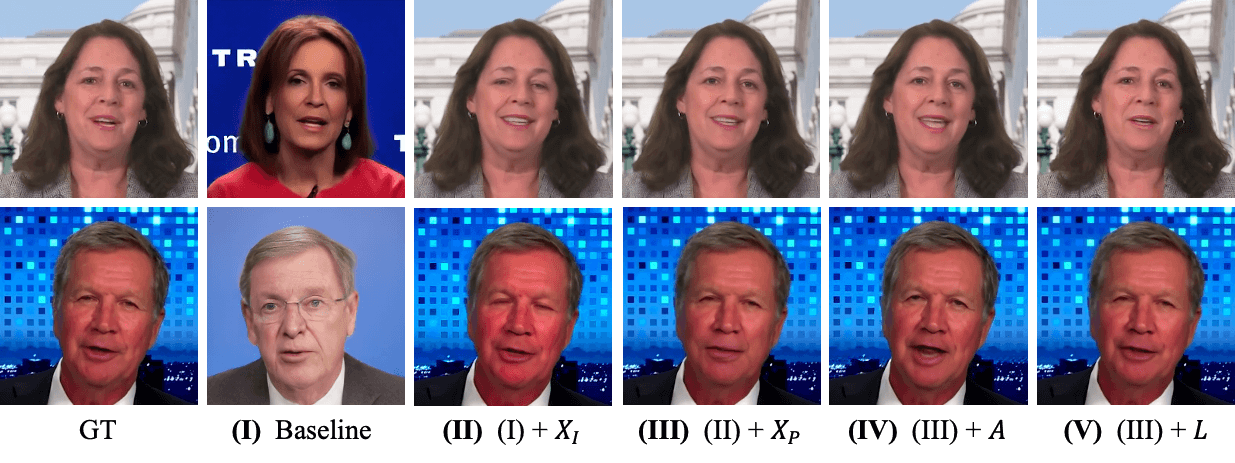}
    \vspace{-20pt}
    \caption{\textbf{Ablation study for MToV.} The baseline model (I), lacking conditions, learns only the general data distribution. Identity frames $X_\mathrm{I}$ as a condition (II) ensure identity preservation throughout the generated video. Additionally, pose frames $X_\mathrm{P}$ in (III) focus generation on the mouth region and improve head pose accuracy. While audio embedding (IV) through cross-attention leads to rough lip movement alignment, facial motion landmarks $L$ (V) significantly enhance lip-sync. 
    }
\label{fig:mcvdm_ablation_qual}
\vspace{-10pt}
\end{figure*}

%% file: assets/sec/5_Conclusion.tex
\section{Conclusion}
In this paper, we present MoDiTalker, a novel motion-disentangled diffusion model for talking head video generation. MoDiTalker addresses the limitations of previous GAN-based methods, including mode collapse and sub-optimal performance, by introducing AToM for high-fidelity lip movement generation from audio and identity inputs.  Additionally, MoDiTalker overcomes the temporal inconsistency and computational cost of prior diffusion-based models through MToV, an efficient video diffusion model using tri-plane representations as conditions. Our model achieves state-of-the-art performance with significantly reduced computational demands compared to earlier diffusion-based efforts. Comprehensive ablation and user studies validate the effectiveness of our approach.

%% file: supple_assets/6_supp.tex
\newpage
\appendix

\begin{center}
\Large
\textbf{Appendix}
\end{center}

In the following, we describe the network architectures and implementation details in Sec.~\ref{implementation}, provide visualized explanations for tri-plane representations in Sec.~\ref{vis_autoencoder}, and include additional experimental results and cross-identity results in Sec.~\ref{additional_results}. We also present a comparison of our model with DreamBooth~\cite{ruiz2023dreambooth} combined with ControlNet~\cite{zhang2023adding} in Sec.~\ref{comp_DB_cont}, detail of the user study in Sec.~\ref{user_study}, limitations in Sec.~\ref{limitations}. 

\section{More implementation details}
\label{implementation}
\subsection{Additional experimental setup for AToM}
\noindent\textbf{Data preparation for training and inference.}
We train AToM to predict the residual between an initial landmark and subsequent audio-synchronized landmarks, conditioned on audio input. Additionally, to ensure AToM predicts the residual focusing more on mouth region without considering head pose, we separate the head pose from facial landmarks. 

For audio input, we extract HuBERT~\cite{hsu2021hubert} features from raw audio, following the process outlined in~\cite{ye2023geneface}. To obtain 3D facial landmarks, we follow these steps: First, we obtain a 3D facial mesh $M$ using Deep 3D Reconstruction~\cite{deng2019accurate}. This technique reconstructs the 3D facial mesh $M$ using a single image. Second, we align the position and pose of $M$ to a frontal orientation, guided by the pose information of the template mean face $\overline{M}$ from~\cite{blanz2023morphable}. This process yields a 3D facial mesh precisely aligned to a frontal view while preserving the person's unique facial features. Finally, we extract 3D facial landmarks from the frontal facial mesh. This process yields audio HuBERT feature and frontal landmark pairs for each video, which we use for both training and inference.

\vspace{5pt}
\noindent\textbf{Network details of AToM.} 
The detailed network architecture of the AToM block, including the audio encoder and landmark encoder, is outlined in Tab.~\ref{tab:AToM_network}. The audio encoder utilizes HuBERT features extracted from raw audio using HuBERT~\cite{hsu2021hubert} and generates audio features that serve as a condition for the AToM. Similarly, the landmark encoder takes an initial landmark as input and produces an initial landmark feature that also acts as a condition for the AToM. The AToM block mainly consists of several transformer decoders and a linear layer. As the sequence passes through a series of AToM blocks, it is designed to predict the residuals of 204 landmark points from 156 frames for each batch, where 204 corresponds to $68\times 3$, indicating the $x$, $y$, and $z$ coordinates of 68 facial landmark points.

\input{supple_assets/tables/tab_AToM_network}

\vspace{5pt}
\noindent\textbf{Implementation details of AToM.} 
We set the input representation dimension to 204, corresponding to 68 landmarks with 3 coordinates for each point ($x$, $y$, and $z$ axes), and fix the latent dimension at 512. We employ the Adan optimizer~\cite{xie2023adan} for this model and adopt a joint loss, which is the weighted sum of the simple reconstruction L2 loss~\cite{zhao2016loss} and the velocity loss~\cite{tevet2022human}. The denoising timestep $T$ is 1000 for sampling. Our model was trained on a single 24GB RTX 3090 for one day with a batch size of 64 and a learning rate of $1e{-4}$.

\newpage

\subsection{Additional experimental setup for MToV}
\noindent\textbf{Data preparation for training and inference.} 
We follow the official documentation~\cite{zhang2021flow} for pre-processing the images of the HDTF dataset. The images were cropped according to the coordinates of the proposed facial regions with a resolution of $256\times256$. For landmark extraction, we used an Deep 3D Reconstruction~\cite{deng2019accurate} that identified 68 facial landmarks with 3D coordinates from the images. These landmarks were then paired with their corresponding images to train our MToV module. 

Since AToM is trained on frontal-aligned facial landmarks to emphasize lip motion over head pose, the position and pose of the generated motion sequence $\hat{L}$ are fixed to those of the mean face $\overline{M}$, respectively. Consequently, generated 3D facial landmarks $\hat{L}$ must be transformed using affine transformations~\cite{wang2021one} to match the face position and pose in the reference video. 
Specifically, 2D landmarks are extracted from each frame of the reference video. Using the extracted 2D landmarks and the generated 3D facial landmarks for each frame, a transformation matrix is calculated.
The calculated transformation matrix are applied to the entire sequence of motion landmarks, aligning them with the face in the reference video. This process is illustrated in Fig.~\ref{fig:Face_align}.

\input{supple_assets/figures/fig_Face_align}

\noindent\textbf{Network details of MToV.} 
The detailed network architectures of the MToV, composed of the landmark autoencoder, pose autoencoder and denoising U-Net, are outlined in Tab.~\ref{tab:AE_config} and Tab.~\ref{tab:unet_config}, respectively. The identity autoencoder shares the same model of pose autoencoder.

For autoenocder, we opt for a 4-layer Transformer. Specific model setting can be found in Tab.~\ref{tab:AE_config}. 

\input{supple_assets/tables/tab_AE_config}

For denoising unet model, we set a learning rate as 1$e$-4 with a batch size of 12. For more hyperparameters, such as the base channel and the depth of U-Net architecture, we adopt a similar setup to LDMs~\cite{rombach2022high} following~\cite{yu2023video}.

\input{supple_assets/tables/tab_Unet_Config}

\noindent\textbf{Implementation details of MToV.}
To obtain the video latent representation, we pass the RGB input video through a video encoder. Similarly, to obtain the reference and identity latent representations, we pass the reference and identity videos through the same video encoder, respectively. In addition, to acquire the motion latent representation as a motion condition, use a motion video encoder on the paired binary motion videos with their corresponding RGB input videos. The hyperparameters and settings used to train the denoising unet are shown in Tab.~\ref{tab:unet_config}.

We trained the autoencoders and video diffusion model separately. For training the autoencoders, we initially trained with a reconstruction loss and then utilized LPIPS~\cite{zhang2018unreasonable} loss to enhance the fidelity of the generated video. The video autoencoder was trained on a single 40GB A100 for 3 days with a batch size of 4 and a learning rate of $4e{-4}$. However, we encountered challenges when using a video encoder that was originally trained on RGB videos to encode motion videos composed of binary values. To overcome this issue, we freeze the decoder in the autoencoder pretrained on RGB videos and subsequently fine-tuned only its specific layer of encoder using motion videos as input data for encoding purposes. 
The video autoencoder can be tuned to a motion video by freezing the decoder and training only the encoder with 1 batch training on a single 24GB 3090 RTX. This process requires 6 hours of additional training time with a learning rate of $1e{-4}$. The reconstruction results can be found in Fig.~\ref{fig:AE_recon_rgb} and Fig.~\ref{fig:AE_recon_ldmk}.
The hyperparameters and settings used to train the video autoencoder and motion video encoder are shown in Tab.~\ref{tab:AE_config}.

The process shown in Fig.~\ref{fig:Tri_plane} converts a video sequence $X$ into a 2D tri-plane representation $Z = \{ z^{hw}, z^{hs}, z^{ws} \}$. This representation decomposes the video's variations along the axes of time $z^{hw}$, height $z^{ws}$, and width $z^{hs}$.

\input{supple_assets/figures/fig_Tri_plane}

\newpage

\section{Visualization of Autoencoder}
\label{vis_autoencoder}
Diffusion models~\cite{rombach2022high, ho2022imagen, ho2022video, yu2023video, hu2023lamd, wang2023leo} have demonstrated superior performance in generating high-quality images and videos based on different conditions (e.g., text, audio). However, since diffusion models perform denoising in the input space, training and inference can be computationally expensive when dealing with high-dimensional data like images and videos. To address this, previous studies~\cite{rombach2022high, ho2022video, yu2023video} have successfully projected high-dimensional data onto low-dimensional latent vectors, performing denoising within the latent space. Following~\cite{yu2023video}, we project video frames consisting of multiple frames into 2D tri-plane representations and train a diffusion model on them.

\newpage
\subsection{Reconstruction visualization of Autoencoder}
In Fig.~\ref{fig:AE_recon_rgb} and Fig.~\ref{fig:AE_recon_ldmk}, we present the reconstruction results $\hat{X}$ for different video types, including RGB videos and binary-valued motion videos, to evaluate the effectiveness of our autoencoder training scheme. 


\input{supple_assets/figures/fig_AE_recon_rgb}

Specifically, after training the autoencoder with RGB video, we freeze the decoder and fine-tune only some layers on top of the pre-trained weights of encoder using the reconstruction loss for binary motion video.

Binary motion video reconstruction results are shown in Fig.~\ref{fig:AE_recon_ldmk}. The first row shows the ground truth (GT), the second row shows the results when the encoder is used without additional fine-tuning, and the third row shows the results when the encoder is used after additional fine-tuning with motion videos. By comparing the results of the second and third rows, we can see that the mouth motion is much better reflected, while the spatial and local information of the motion video is better preserved. This means that a latent containing more accurate motion information can be used as a condition.

\input{supple_assets/figures/fig_AE_recon_ldmk}

\newpage
\subsection{Latents visualization of Autoencoder}
To show what each tri-plane represents, we visualize the intermediate latents of each $z^{hw}, z^{hs}$ and $z^{ws}$ planes.

A single video frame composed of multiple frames can be projected onto three 2D planes. Fig.~\ref{fig:AE_feature} shows the visualization results of the latents for three video frames of three different people. The leftmost three columns show the visualization results of the latents for the GT video frame, middle three columns show the visualization results of the latents for the motion video frame, and rightmost three columns show the visualization results of the denoised $z_0$.

\input{supple_assets/figures/fig_AE_feature}

\newpage
\newpage
\section{Additional results}
\label{additional_results}
\subsection{Cross identity setting including GAN-based methods}
We conducted an experiment under the cross-identity setting, where a single audio and different identity conditions are given to generate talking head videos in Fig.~\ref{fig:Cross_id}. By incorporating the process of generating intermediate motion from the audio, we are able to effectively separate the correlation between the audio and the generated speaker.

\input{supple_assets/figures/fig_Cross_id}

\newpage
\subsection{One-shot pose frame setting}
We can also create a talking head video in a setting where only a one-shot image is given in Fig.~\ref{fig:One_shot}.

\input{supple_assets/figures/fig_One_shot}

\newpage
\subsection{Unseen dataset}
AToM is trained solely on the LRS3-TED dataset~\cite{afouras2018lrs3}. To demonstrate its capabilities on unseen data, we validate AToM using HDTF~\cite{zhang2021flow} audio in the main paper. We focus on the HDTF dataset for MToV training due to its high-quality images. To demonstrate MToV's capabilities on unseen data, we also present results from the widely used the CREMA-D dataset~\cite{cao2014crema} and VoxCeleb2 dataset~\cite{chung2018voxceleb2} in Fig.~\ref{fig:Crema} and Fig.~\ref{fig:Vox}.

\vspace{5pt}
\noindent\textbf{CREMA.}
We apply the same audio and frame processing techniques used for HDTF to the CREMA-D dataset. For consistency with our model's training on HDTF's $256\times256$ resolution, we resize images of CREMA-D from $320\times320$ to $256\times256$. The results are shown in Fig.~\ref{fig:Crema}.

\input{supple_assets/figures/fig_Crema}

\newpage
\noindent\textbf{VoxCeleb2.}
We apply the same audio and frame processing techniques used for HDTF to the VoxCeleb2 dataset. To align with the resolution of HDTF employed in the training, we resize the images of VoxCeleb2 from $224\times224$ to $256\times256$. The results are shown in Fig.~\ref{fig:Vox}.
\input{supple_assets/figures/fig_Vox}

\newpage
\noindent\textbf{Diffused Heads on CREMA.}
We confirmed that Diffused Heads~\cite{stypulkowski2024diffused} perform poorly with HDTF. This is because Diffused Heads only release pre-trained weights for CREMA-D and do not provide any training code. To demonstrate that we properly evaluated the Diffused Heads code, we compared the results of Diffused Heads and MoDiTalker. MoDiTalker is trained with HDTF, so CREMA-D is unseen dataset.

\input{supple_assets/figures/fig_Crema_diffused}

\newpage
\section{Comparison with DreamBooth \& ControlNet}
\label{comp_DB_cont}
As shown in Fig.~\ref{fig:DreamBooth_ControlNet}, we also compared the results with general-purpose conditional diffusion models, not specifically designed for the talking head task.

\input{supple_assets/figures/fig_DreamBooth_ControlNet}
DreamBooth~\cite{ruiz2023dreambooth} is a method for personalizing people, objects, and styles with Stable Diffusion (SD) models trained on large datasets. ControlNet~\cite{zhang2023adding} has become popular as an additional model for generating images by conditioning on pretrained SD. By using DreamBooth to optimize SD for a specific person and ControlNet tuned to a facial motion dataset together, we can create a talking head video personalized for that individual as shown in Fig.~\ref{fig:DreamBooth_ControlNet}.

\newpage

\section{User study details}
\label{user_study}
We conducted a user study to compare MoDiTalker against prior GAN-based~\cite{prajwal2020lip, zhou2021pose, zhou2020makelttalk, wang2021audio2head} and diffusion-based~\cite{stypułkowski2023diffused, shen2023difftalk, ma2023dreamtalk} approaches. 

Participants evaluated 15 randomly selected 5-second segments generated using HDTF audio and identity frames. Our study focused on lip-sync fidelity, identity preservation, and video quality. For lip-sync fidelity, participants were provided with audio input and generated talking head videos from different methods, and were asked, `Which video do you think has the best lip synchronization?' For identity preservation, we provided an identity frame and generated talking head videos from various methods, and asked participants, `Which video do you think preserves identity the most?' Lastly, for video quality, we presented generated talking head videos from different studies and asked participants, `Which video do you think has the highest quality?' 

Results shown in main paper demonstrate that MoDiTalker surpasses previous works in all aspects, highlighting the effectiveness of our approach.

\section{Limitations}
\label{limitations}
Although MoDiTalker demonstrates valuable performance in various settings, our model inherits some limitations. While our model creates video frames by generating multiple frames simultaneously, ensuring temporal consistency within frames, there are instances where temporal consistency between frames is less ideal. This issue could be alleviated with additional post-processing. Additionally, the lack of dynamic poses in the HDTF dataset poses a challenge in generating dynamic and diverse poses, reflecting constraints within the dataset itself.

%% file: supple_assets/tables/tab_AToM_network.tex
\begin{table}[ht] 
\centering
\small{
 \scalebox{0.9}{
        \begin{tabular}{>{\centering}m{0.3\linewidth} >{\centering}m{0.1\linewidth} 
                >{\centering}m{0.25\linewidth}>{\centering}m{0.30\linewidth}}
                \textbf{Audio Encoder}&&& \tabularnewline
            \hline
            Layer& w.FiLM & Norm & Output shape $(C \times H \times W)$ \tabularnewline
            \hline
            Transformer Encoder& - & LayerNorm & $(64,312,512)$  \tabularnewline
            Transformer Encoder& - &LayerNorm  & $(64,312,512)$   
            \tabularnewline
            \hline
            &&\tabularnewline 
            \textbf{Landmark Encoder}&&& \tabularnewline \hline
            Layer& w.FiLM & Norm & Output shape $(C\times H \times W)$  \tabularnewline\hline
            Transformer Encoder & - &LayerNorm & $(64,156,512)$  \tabularnewline
            Transformer Encoder & - &LayerNorm  & $(64,156,512)$   \tabularnewline

            \hline
            &&\tabularnewline 
            \textbf{AToM Block}&&& \tabularnewline \hline
            Layer & w.FiLM & Norm & Output shape $(C \times H \times W)$ \tabularnewline
            \hline
            Transformer Decoder & O & LayerNorm & $(64, 156, 1024)$  \tabularnewline
            Transformer Decoder & O & LayerNorm & $(64, 156, 1024)$  \tabularnewline
            Transformer Decoder & O & LayerNorm & $(64, 156, 1024)$  \tabularnewline
            Transformer Decoder & O & LayerNorm & $(64, 156, 1024)$  \tabularnewline
            Transformer Decoder & O & LayerNorm & $(64, 156, 1024)$  \tabularnewline
            Transformer Decoder & O & LayerNorm & $(64, 156, 1024)$  \tabularnewline
            Transformer Decoder & O & LayerNorm & $(64, 156, 1024)$  \tabularnewline
            Transformer Decoder & O & LayerNorm & $(64, 156, 1024)$  \tabularnewline
            Linear & - & - & $(64, 156, 204)$  \tabularnewline
            
            \hline
        \end{tabular}}
        \vspace{5pt}
        \caption{\textbf{Network architecture of AToM.}}
        \label{tab:AToM_network}
    }
\end{table}

%% file: supple_assets/figures/fig_Face_align.tex
\begin{figure}[h]
  \centering
  \includegraphics[width=1\linewidth]{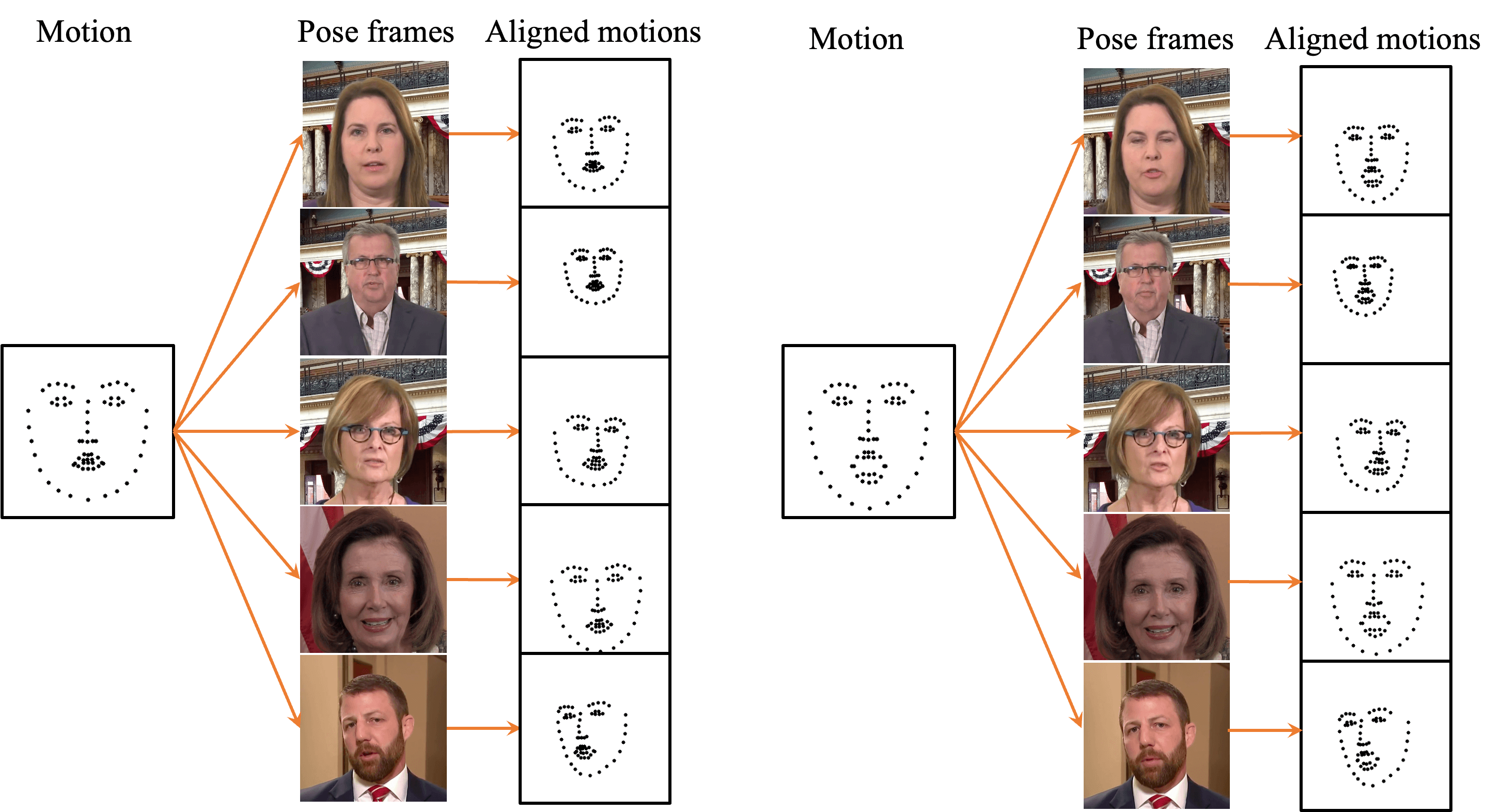}
  \vspace{-10pt}
  \caption{\textbf{Motion aligning.} The facial motion, which is the output of AToM with a position and pose fixed to the mean face of 3DMM~\cite{blanz2023morphable}, is converted to a facial motion with a pose aligned to each frame of the reference video by extracting transformation matrix from the each reference frame.}
\label{fig:Face_align}
\end{figure}

%% file: supple_assets/tables/tab_AE_config.tex
\vspace{-5pt}
\begin{table}[h]
    \centering
    \resizebox{\linewidth}{!}{
    \begin{tabular}{l cccccc}
    \toprule
         & {Channel} 
         & {\# ResBlock} 
         & {Timesteps} 
         & {Input res.} 
         & {Emb. dim.} 
         & {\# Iter.} \\
         
    \midrule
    {pose autoencoder} 
    & 384 & 2 & 16 & 256 & 4 & 400k \\
    {landmark autoencoder} 
    & 384 & 2 & 16 & 256 & 4 & 60k \\
         
    \bottomrule
    \end{tabular}
    }
    \vspace{5pt}
    \caption{\textbf{Training settings for pose video autoencoder and motion video encoder.}}
    \vspace{-20pt}
    \label{tab:AE_config}
\end{table}

%% file: supple_assets/tables/tab_Unet_Config.tex
\begin{table}[h]
    \centering
    \resizebox{\linewidth}{!}{
    \begin{tabular}{l ccccccccc}
    \toprule
         & {Base channel} 
         & {Attn. res.} 
         & {\# ResBlock} 
         & {Channel mul.} 
         & {\# Heads} 
         & {Linear start} 
         & {Linear end} \\
         
    \midrule
    {denoising unet} & 128 & {[4,2,1]} & 2 & {[1,2,4,4]} & 8 & 0.0015 & 0.0195  \\
    
    \bottomrule
    \end{tabular}
    }
    \vspace{5pt}
    \caption{\textbf{Training settings for denoising unet.}}
    \vspace{-10pt}
    \label{tab:unet_config}
\end{table}

%% file: supple_assets/figures/fig_Tri_plane.tex
\begin{figure}[h]
  \centering
  \includegraphics[width=1\linewidth]{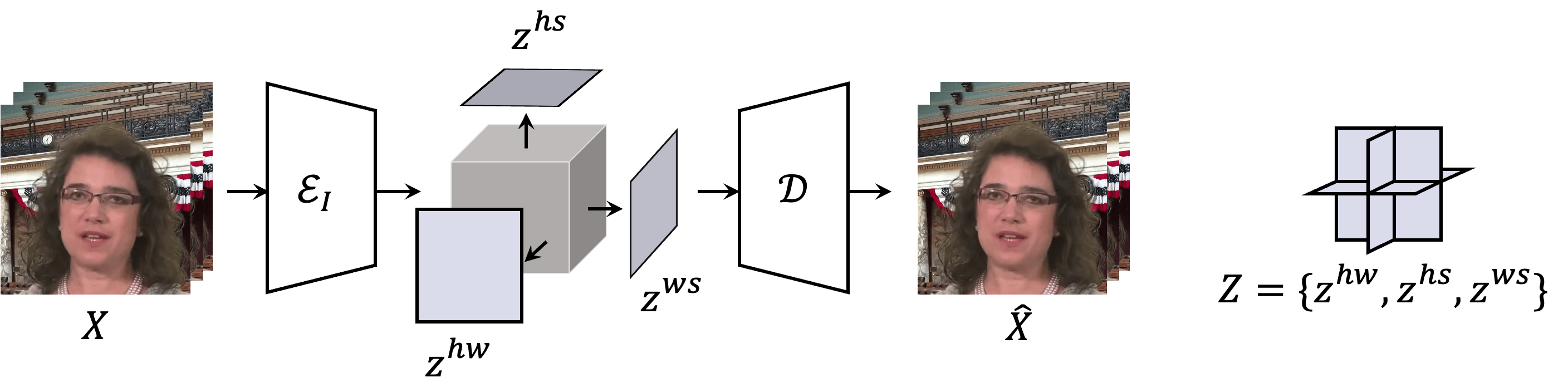}
  \vspace{-13pt}
    \caption{\textbf{Extracting tri-plane representations from video frames.} We project video frames into tri-plane representations, decomposing the video along the axes of time, height, and width.}
\label{fig:Tri_plane}
\end{figure}

%% file: supple_assets/figures/fig_AE_recon_rgb.tex
\begin{figure}[h]
  \centering
  \includegraphics[width=1\linewidth]{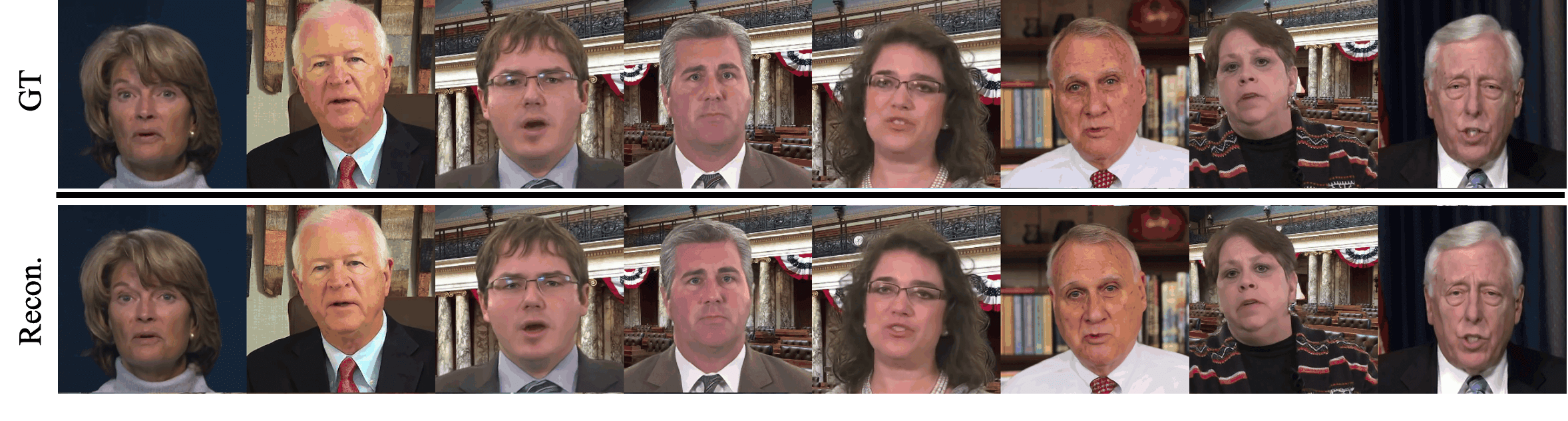}
  \vspace{-20pt}
  \caption{\textbf{Reconstruction results of pose video frames.}}
  \vspace{-10pt}
\label{fig:AE_recon_rgb}
\end{figure}

%% file: supple_assets/figures/fig_AE_recon_ldmk.tex
\begin{figure}[h]
  \centering
  \includegraphics[width=1\linewidth]{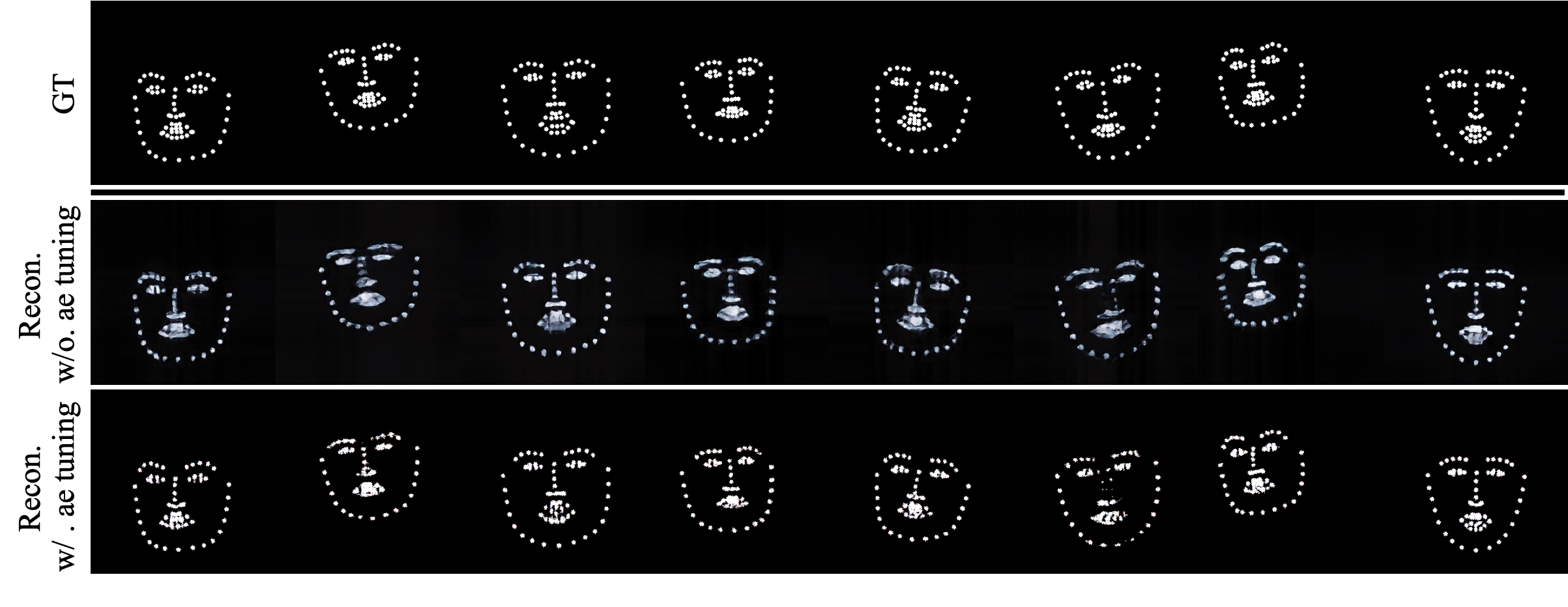}
  \vspace{-20pt}
  \caption{\textbf{Reconstruction results of motion video frames.}}
  \vspace{-10pt}
\label{fig:AE_recon_ldmk}
\end{figure}

%% file: supple_assets/figures/fig_AE_feature.tex
\begin{figure}[h]
  \centering
  \includegraphics[width=1\linewidth]{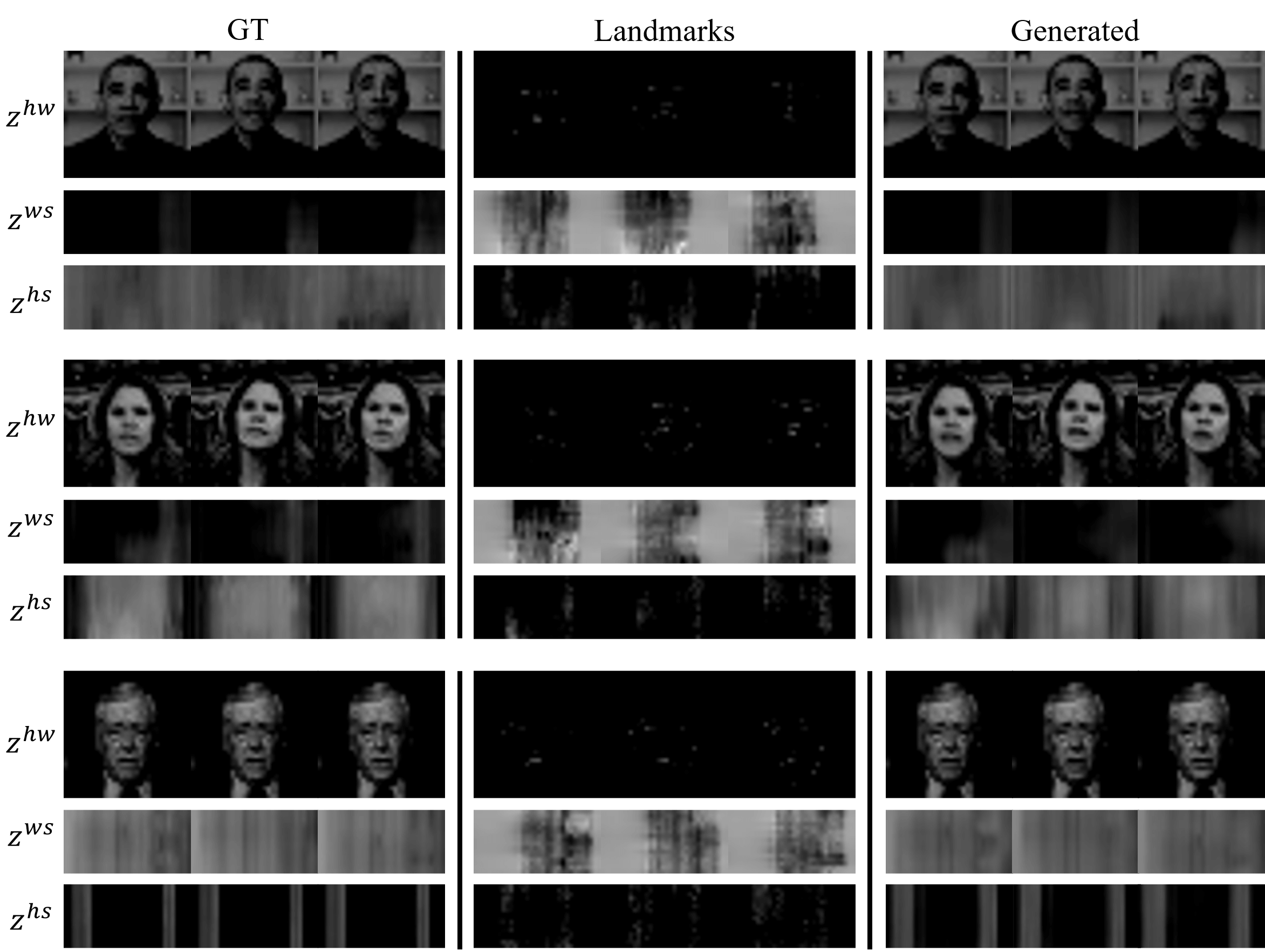}
  \vspace{-15pt}
  \caption{\textbf{Latents visualization of autoencoder}.}
\label{fig:AE_feature}
\end{figure}

%% file: supple_assets/figures/fig_Cross_id.tex
\begin{figure}[h]
  \centering
  \includegraphics[width=1\linewidth]{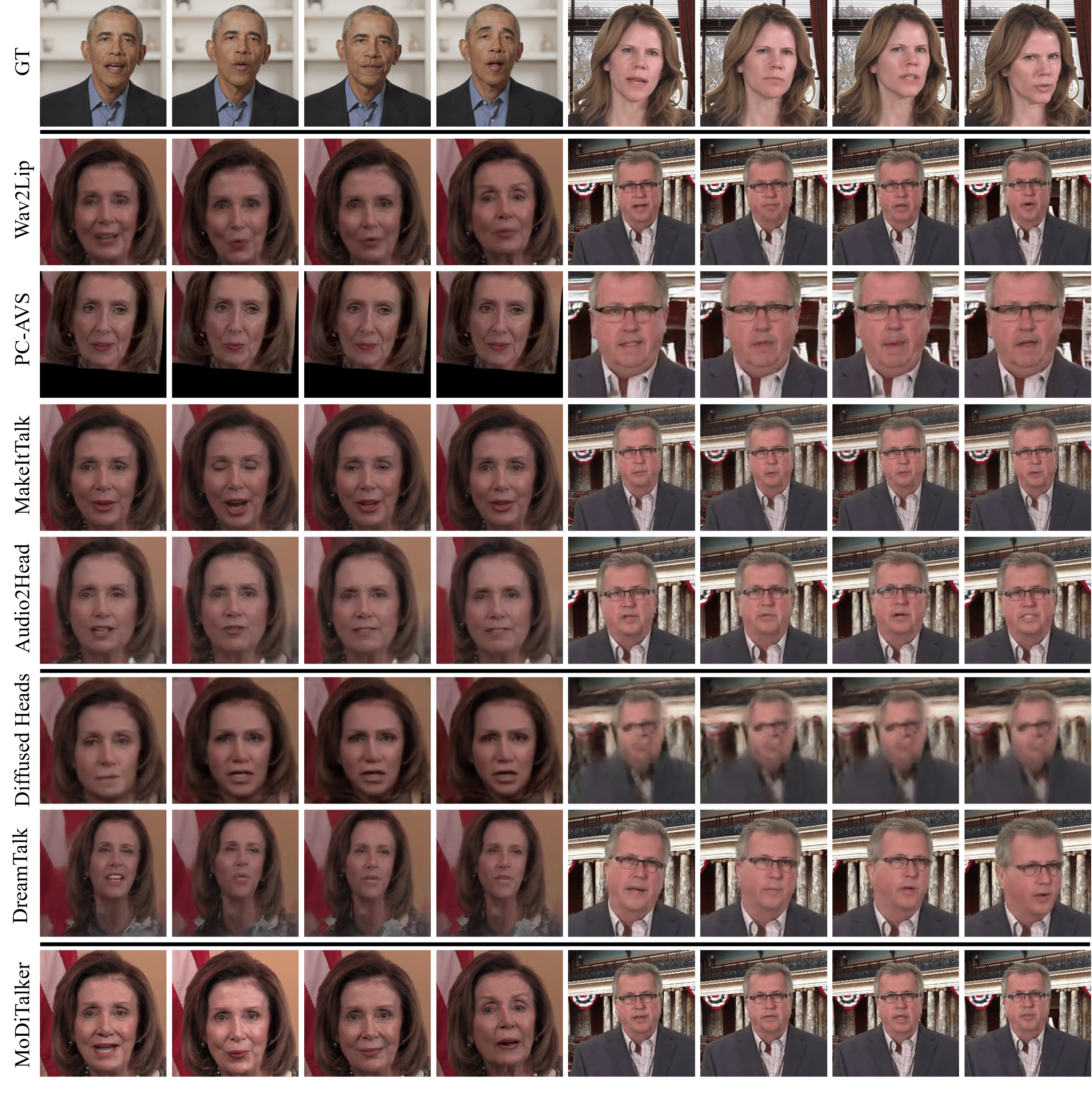}
  \vspace{-17pt}
  \caption{\textbf{Qualitative comparison with previous works on cross identity setting.} We compare MoDiTalker with previous GAN-based methods~\cite{prajwal2020lip, zhou2021pose, zhou2020makelttalk, wang2021audio2head}, including Wav2Lip, PC-AVS, MakeItTalk, Audio2Head and diffusion-based methods~\cite{stypulkowski2024diffused, ma2023dreamtalk}, including Diffused Heads, DreamTalk.}
\label{fig:Cross_id}
\end{figure}

%% file: supple_assets/figures/fig_One_shot.tex
\begin{figure}[h]
  \centering
  \includegraphics[width=0.8\linewidth]{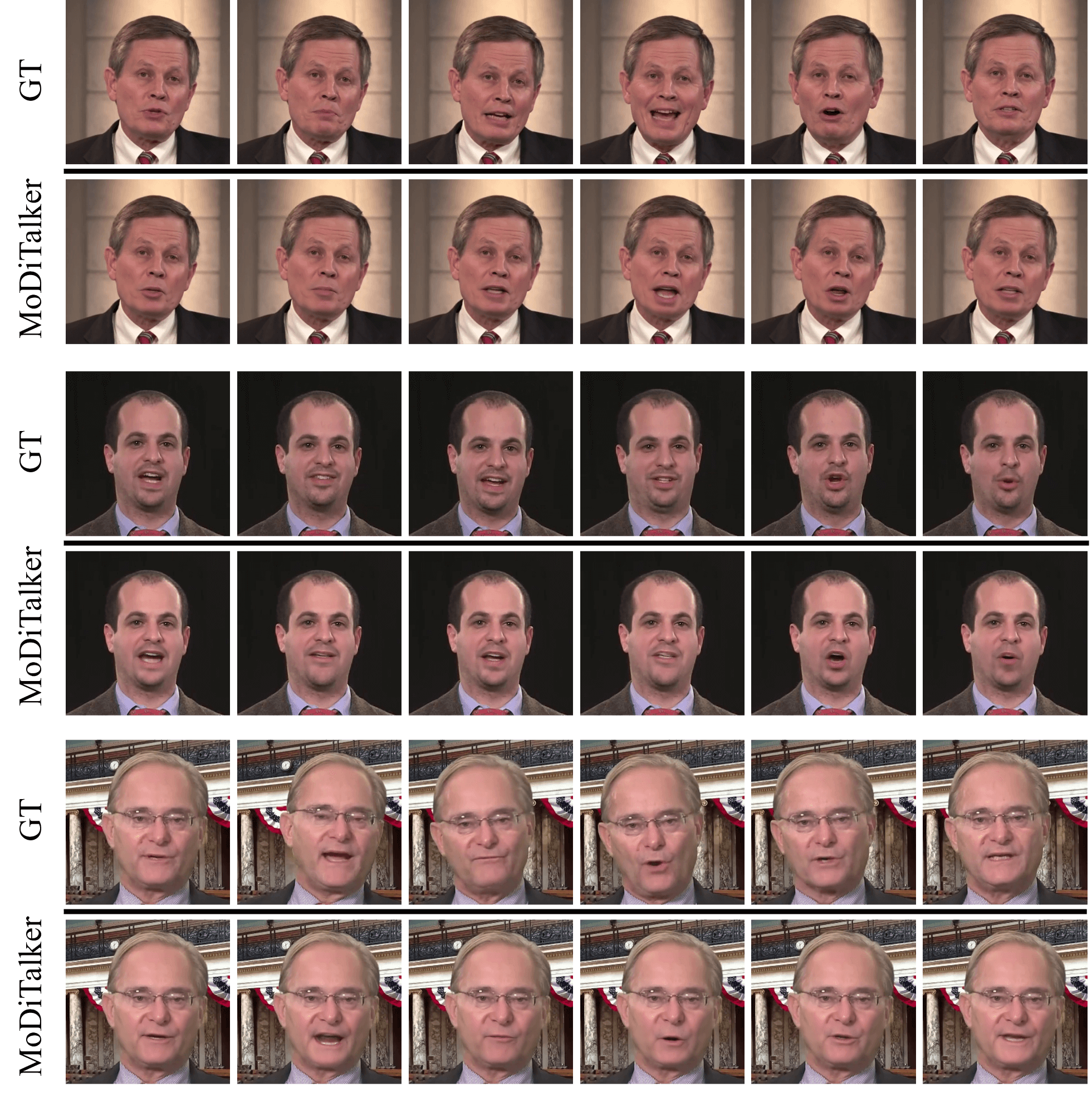}
  \vspace{-5pt}
  \caption{\textbf{Qualitative results for oneshot frame setting.} }
\label{fig:One_shot}
\end{figure}

%% file: supple_assets/figures/fig_Crema.tex
\begin{figure}[h]
  \centering
  \includegraphics[width=0.8\linewidth]{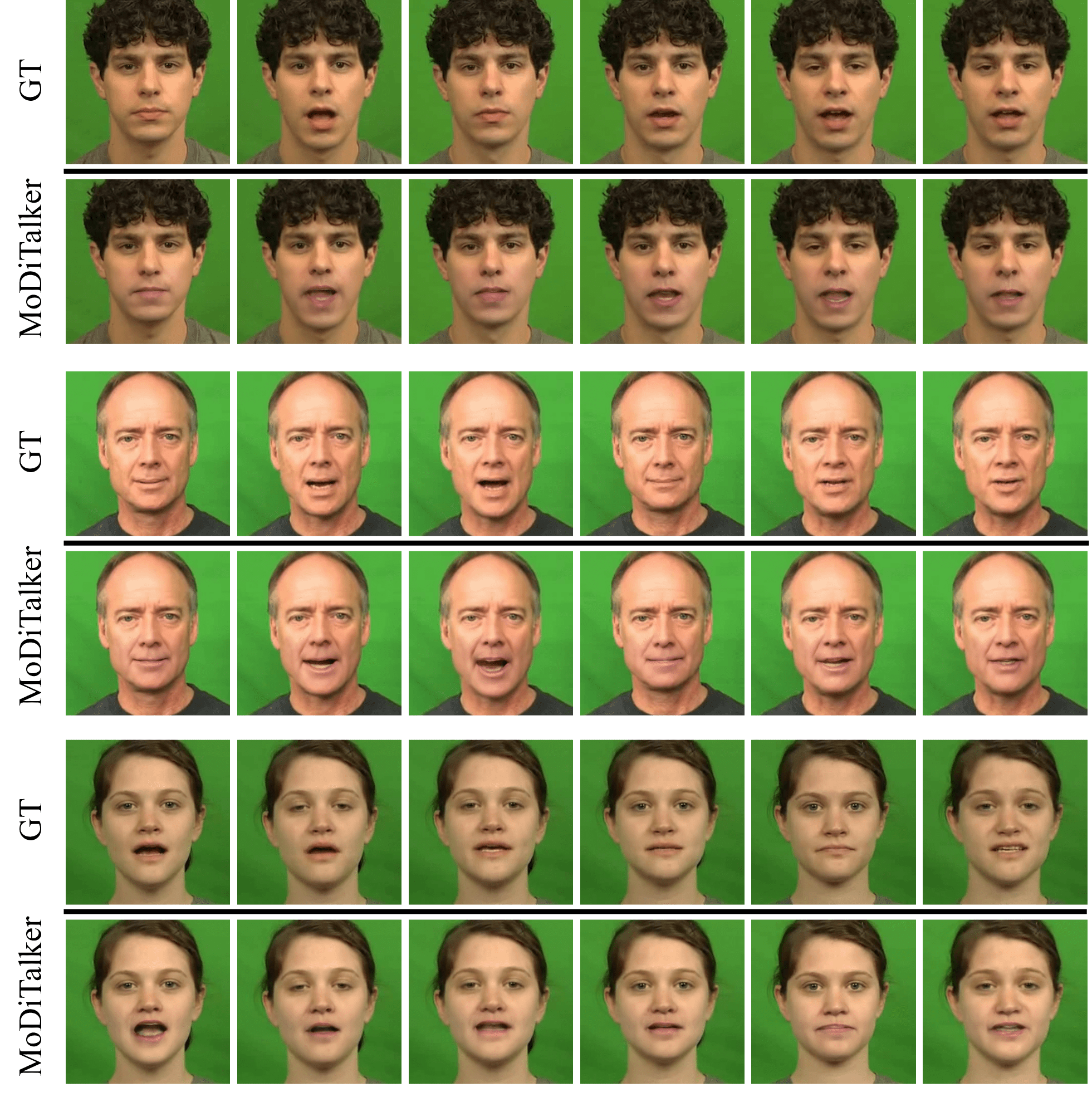}
  \vspace{-5pt}
  \caption{\textbf{Qualitative results for unseen CREMA~\cite{cao2014crema} dataset setting.}}
\label{fig:Crema}
\end{figure}

%% file: supple_assets/figures/fig_Vox.tex
\begin{figure}[h]
  \centering
  \includegraphics[width=0.8\linewidth]{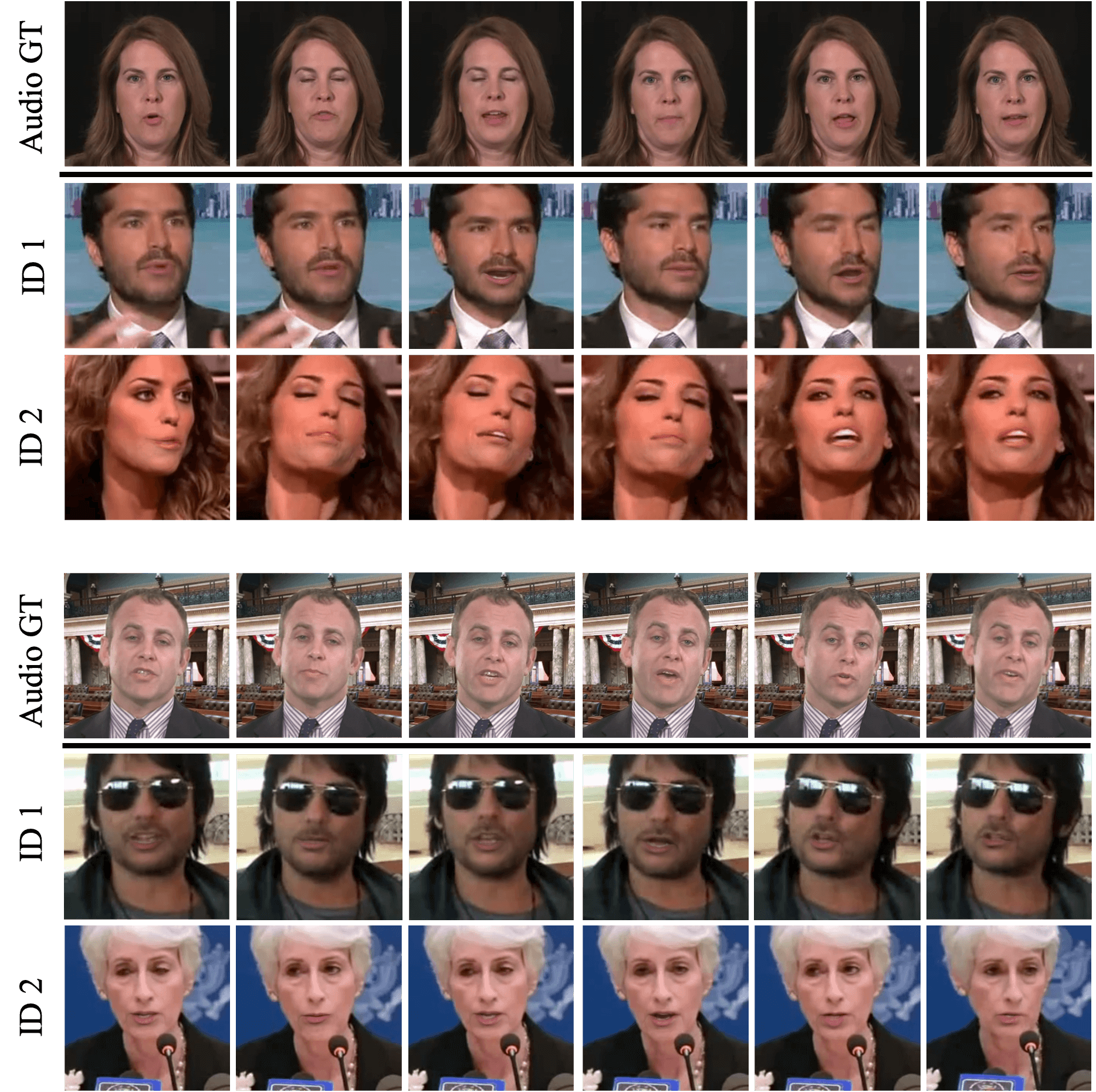}
  \vspace{-5pt}
  \caption{\textbf{Qualitative results for unseen VoxCeleb2 dataset setting.}}
\label{fig:Vox}
\end{figure}

%% file: supple_assets/figures/fig_Crema_diffused.tex
\begin{figure}[h]
  \centering
  \includegraphics[width=0.8\linewidth]{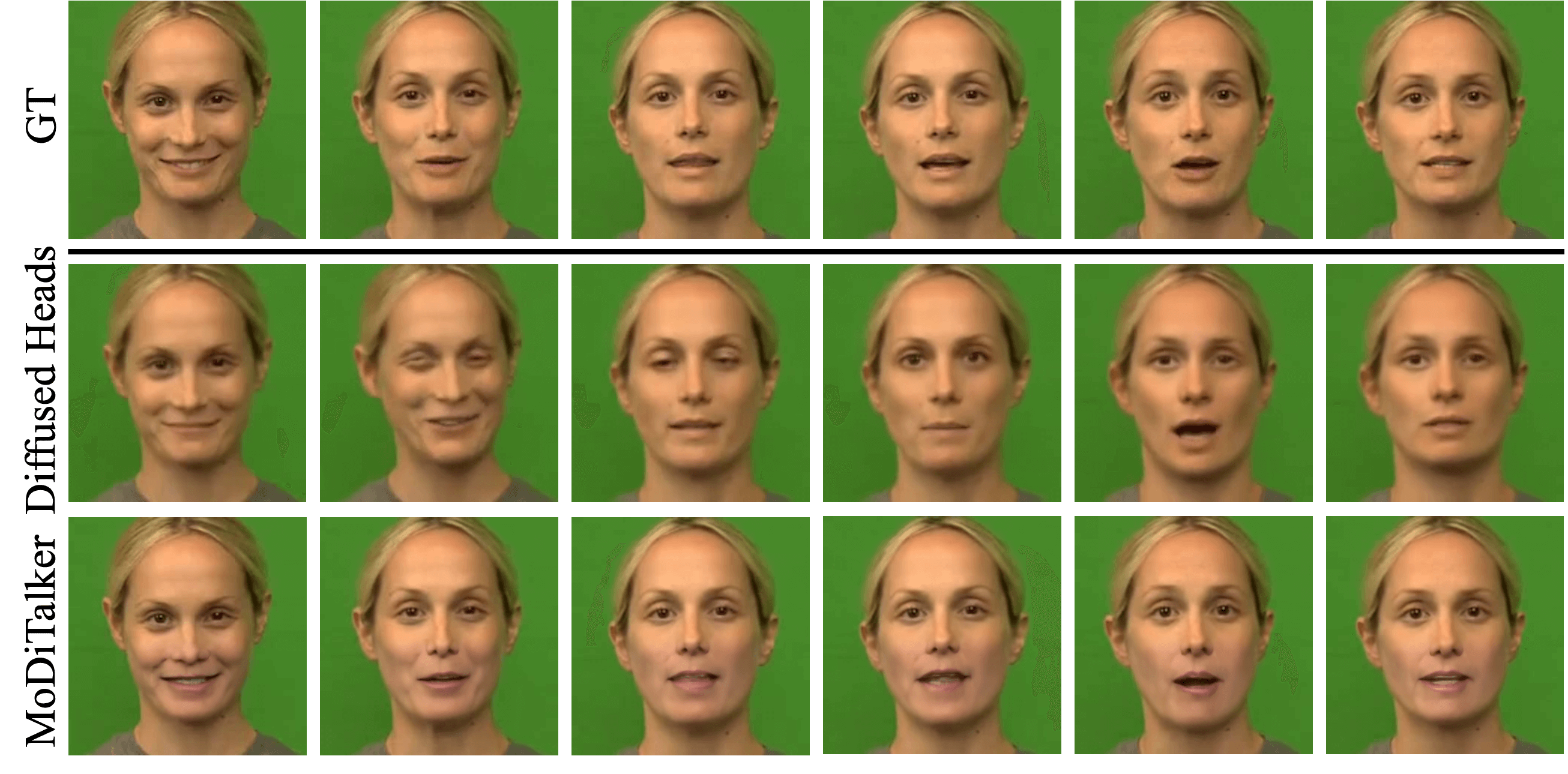}
  \vspace{-5pt}
  \caption{\textbf{Qualitative comparison between Diffused Heads~\cite{stypulkowski2024diffused} and MoDiTalker for CREMA~\cite{cao2014crema} dataset.}}
\label{fig:Crema_diffused}
\end{figure}

%% file: supple_assets/figures/fig_DreamBooth_ControlNet.tex
\begin{figure}[h]
  \centering
  \includegraphics[width=0.8\linewidth]{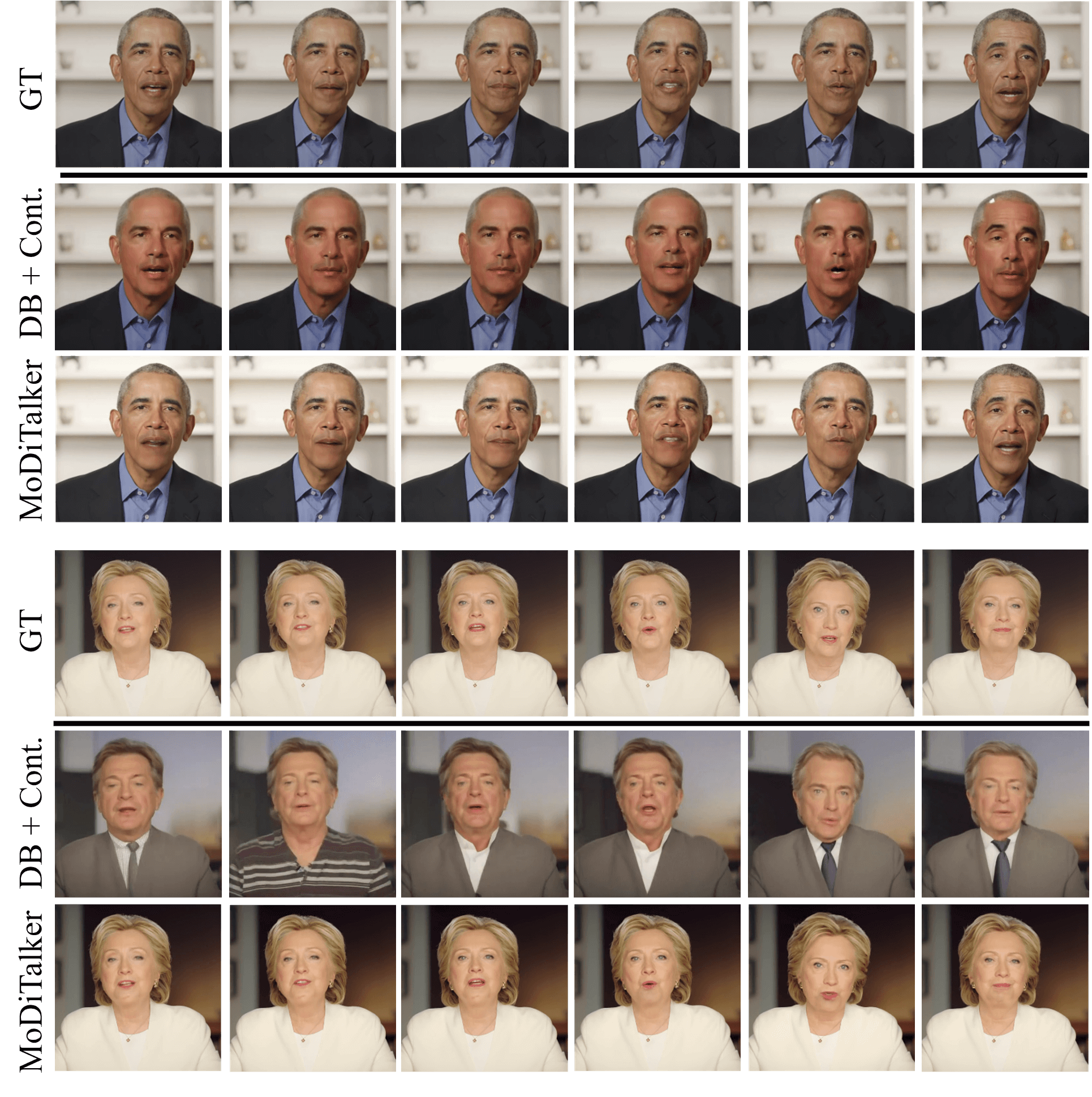}
  \vspace{-10pt}
  \caption{\textbf{Qualitative comparison with DreamBooth (DB)~\cite{ruiz2023dreambooth} and ControlNet (Cont.)~\cite{zhang2023adding}.}}
\label{fig:DreamBooth_ControlNet}
\end{figure}